\documentclass[lettersize,journal]{IEEEtran}
\usepackage{amsmath,amsfonts}
\usepackage{algorithm,algpseudocode}
\usepackage{array}
\usepackage[caption=false,font=normalsize,labelfont=sf,textfont=sf]{subfig}
\usepackage{textcomp}
\usepackage{stfloats}
\usepackage{orcidlink}
\usepackage{url}
\usepackage{multicol}
\usepackage{multirow}
\usepackage{verbatim}
\usepackage{graphicx}
\usepackage{cite}
\usepackage{algorithm,algpseudocode}
\usepackage[dvipsnames]{xcolor}
\usepackage{comment}
\usepackage{xcolor}
\usepackage{hyperref}
\usepackage{booktabs}
\usepackage[export]{adjustbox}
\usepackage[table]{xcolor}
\usepackage{amssymb}
\usepackage{pifont}
\newcommand{\cmark}{\ding{51}}%
\newcommand{\xmark}{\ding{55}}%
\definecolor{skyblue}{HTML}{6ecff6}
\definecolor{lavenderpink}{HTML}{e8b2e8}
\definecolor{leafgreen}{HTML}{44bf55}
\begin{document}

\title{MSF-Mamba: Motion-aware State Fusion Mamba for Efficient Micro-Gesture Recognition}
\author{Deng Li~\orcidlink{0000-0002-7140-0701}, Jun Shao~\orcidlink{0009-0001-4316-6890}, Bohao Xing~\orcidlink{0009-0005-5924-4178}, Rong Gao~\orcidlink{0009-0007-8938-8222}, Bihan Wen~\orcidlink{0000-0002-6874-6453},~\IEEEmembership{Senior member,~IEEE}, Heikki Kälviäinen~\orcidlink{0000-0002-0790-6847}~\IEEEmembership{Senior member,~IEEE}, Xin Liu$\dagger$~\orcidlink{0000-0002-2242-6139},~\IEEEmembership{Senior member,~IEEE}
\thanks{$\dagger$Corresponding author: Xin Liu (email: linuxsino@gmail.com). This work was supported
in part by the China Scholarship Council (Grant
No.~202307960007, No.~202406250043 and No.202306250014). The authors wish to acknowledge CSC – IT Center for Science, Finland, for computational resources.}
\thanks{Deng Li, Bohao Xing, Rong Gao, and Xin Liu are with the Computer Vision and Pattern Recognition Laboratory, Department of Computational Engineering, Lappeenranta-Lahti University of Technology LUT, Lappeenranta, Finland (e-mail: deng.li@lut.fi, bohao.xing@lut.fi, rong.gao@lut.fi, and linuxsino@gmail.com).}
\thanks{Jun Shao is with the School of Electrical and Information Engineering, Tianjin University, Tianjin, China (e-mail: shaojun1017@tju.edu.cn).}
\thanks{Bihan Wen is with the School of Electrical and Electronic Engineering, Nanyang Technological University, Singapore (bihan.wen@ntu.edu.sg).}
\thanks{Heikki Kälviäinen is with the Computer Vision and Pattern Recognition Laboratory, Department of Computational Engineering, Lappeenranta-Lahti University of Technology LUT, Lappeenranta, Finland, and Faculty of Information Technology,  Brno University of Technology (BUT), Brno, Czech Republic (heikki.kalviainen@lut.fi)}
}

\markboth{IEEE Transactions on Multimedia}%
{Shell \MakeLowercase{\textit{et al.}}: A Sample Article Using IEEEtran.cls for IEEE Journals}


\maketitle

\begin{abstract}
Micro-gesture recognition (MGR) targets the identification of subtle and fine-grained human motions and requires accurate modeling of both long-range and local spatiotemporal dependencies. While convolutional neural networks (CNNs) are effective at capturing local patterns, they struggle with long-range dependencies due to their limited receptive fields. Transformer-based models address this limitation through self-attention mechanisms but suffer from high computational costs. \textcolor{black}{Recently, Mamba has shown promise as an efficient model, leveraging state space models (SSMs) to enable linear-time processing.} However, directly applying the vanilla Mamba to MGR may not be optimal. This is because Mamba processes inputs as 1D sequences, with state updates relying solely on the previous state, and thus lacks the ability to model local spatiotemporal dependencies. In addition, previous methods lack a design of motion-awareness, which is crucial in MGR. To overcome these limitations, we propose motion-aware state fusion mamba (MSF-Mamba), which enhances Mamba with local spatiotemporal modeling by fusing local contextual neighboring states. Our design introduces a motion-aware state fusion module based on central frame difference (CFD). Furthermore, \textcolor{black}{a multiscale version} named MSF-Mamba$^{+}$ has been proposed. Specifically, MSF-Mamba$^{+}$ supports multiscale motion-aware state fusion, as well as an adaptive scale weighting module that dynamically weighs the fused states across different scales. These enhancements explicitly address the limitations of vanilla Mamba by enabling motion-aware local spatiotemporal modeling, allowing MSF-Mamba and MSF-Mamba$^{+}$ to effectively capture subtle motion cues for MGR. Experiments on two public MGR datasets (i.e., SMG and iMiGUE) demonstrate that even the lightweight version, namely, MSF-Mamba, achieves state-of-the-art performance, outperforming existing CNN-, Transformer-, and SSM-based models while maintaining high efficiency. For example, MSF-Mamba improves Top-1 accuracy by +2.2\% and +1.5\% over VideoMamba on SMG and iMiGUE, respectively. MSF-Mamba$^{+}$ outperforms VideoMamba on SMG and iMiGUE, achieving Top-1 accuracy improvements of 2.9\% and 3.0\%, respectively. \textcolor{black}{The code will be accessible on \url{https://github.com/Leedeng/MSF-Mamba}}
\end{abstract}
\begin{IEEEkeywords}
Micro Gesture Recognition, Mamba, State Space Model
\end{IEEEkeywords}

\section{Introduction}
Micro-gesture recognition (MGR)~\cite{gu2025motion,li2024mmad} is a key task in computer vision and video analysis and plays an important role in applications such as human-computer interaction~\cite{gupta2023survey,kandoi2023intentional,sinha2010human,li2025repetitive} and emotion understanding~\cite{xing2024emo,gao2024identity,li2024eald,10518114,li2025deemo}. MGR focuses on identifying subtle and fine-grained micro-gestures (MGs)~\cite{chen2019analyze,gong2025survey}, such as \textit{"crossing fingers"} and \textit{"touching jaw"}. In MGR~\cite{chen2024prototype,li2023joint}, capturing fine-grained local dependencies is essential for detecting subtle motion patterns, whereas modeling global dependencies provides crucial contextual information. Over the past few years, deep learning-based methods have shown promising results in action recognition~\cite{sun2022human,wu2024ensemble}. Convolutional neural networks (CNNs)~\cite{lecun2015deep} have been widely adopted due to their ability to capture local spatiotemporal features. These methods~\cite{guo2024benchmarking,lin2019tsm,carreira2017quo,tran2015learning} process video frames using fixed-size receptive fields~\cite{karpathy2014large}, which makes them effective for modeling localized motion patterns~\cite{feichtenhofer2019slowfast}. \textcolor{black}{However, the limited receptive fields inherent to CNNs often hinder their ability to capture global context~\cite{noman2024elgc}.} Vision Transformer~\cite{dosovitskiy2020image} has demonstrated strong capabilities in modeling global dependencies across spatial and temporal dimensions through self-attention mechanisms~\cite{vaswani2017attention}. Recent Transformer-based action recognition methods~\cite{bertasius2021space,liu2022video,li2023uniformerv2,ShuLearning2024,li2023data,liu2025online} further incorporate local context via hierarchical or windowed attention schemes. Despite their effectiveness, Transformers suffer from high computational complexity, namely, scaling quadratically with the input sequence length~\cite{jiao2023dilateformer}. This poses a limitation for scenarios involving real-time or lightweight MGR deployments.

\begin{figure*}[t]
    \centering
    \footnotesize
    \begin{tabular}{cc}
         \includegraphics[width=0.45\linewidth]{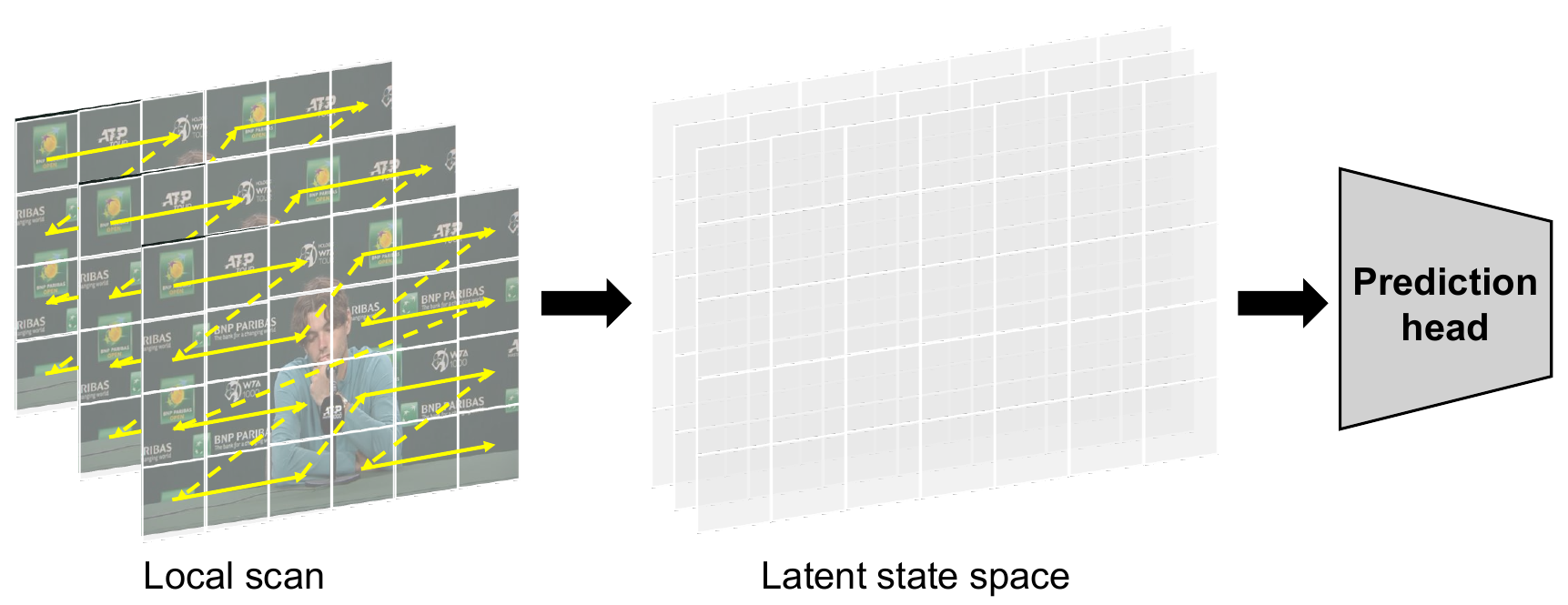}&\includegraphics[width=0.45\linewidth]{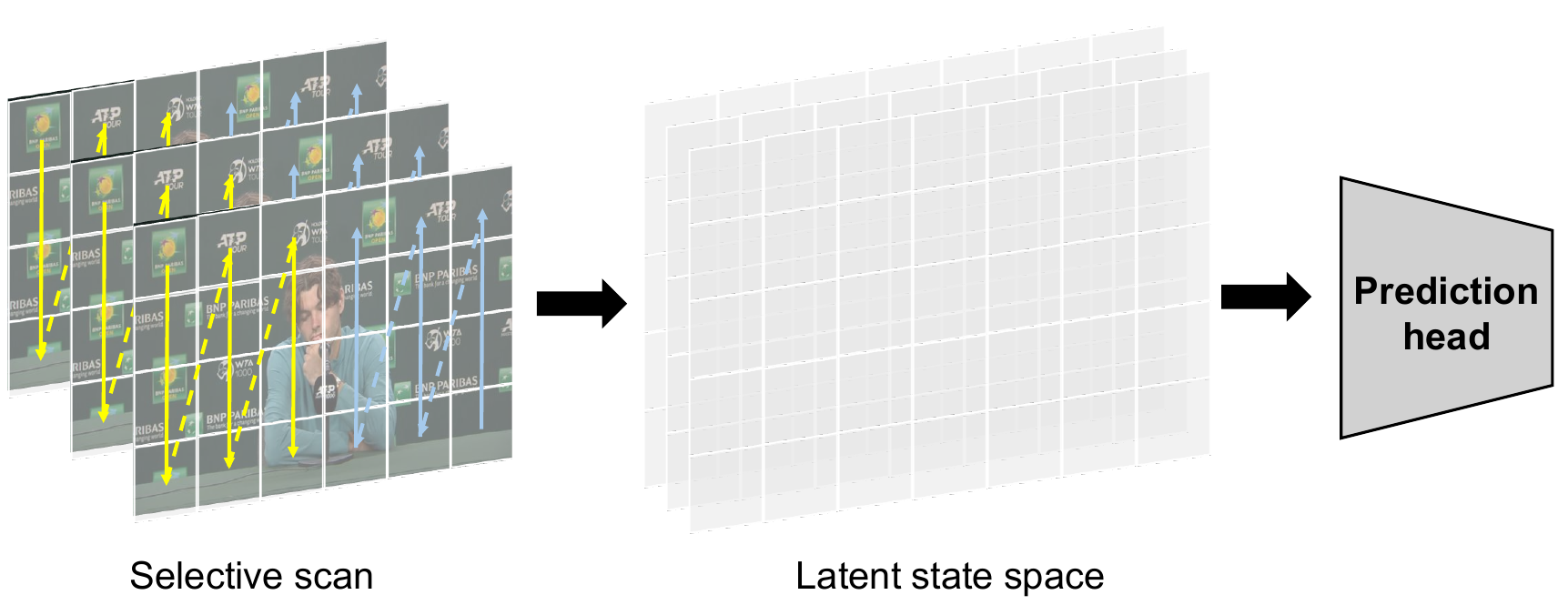}  \\
         (a) Local scan.&(b) Selective scan.\\
         \multicolumn{2}{c}{\includegraphics[width=0.9\linewidth]{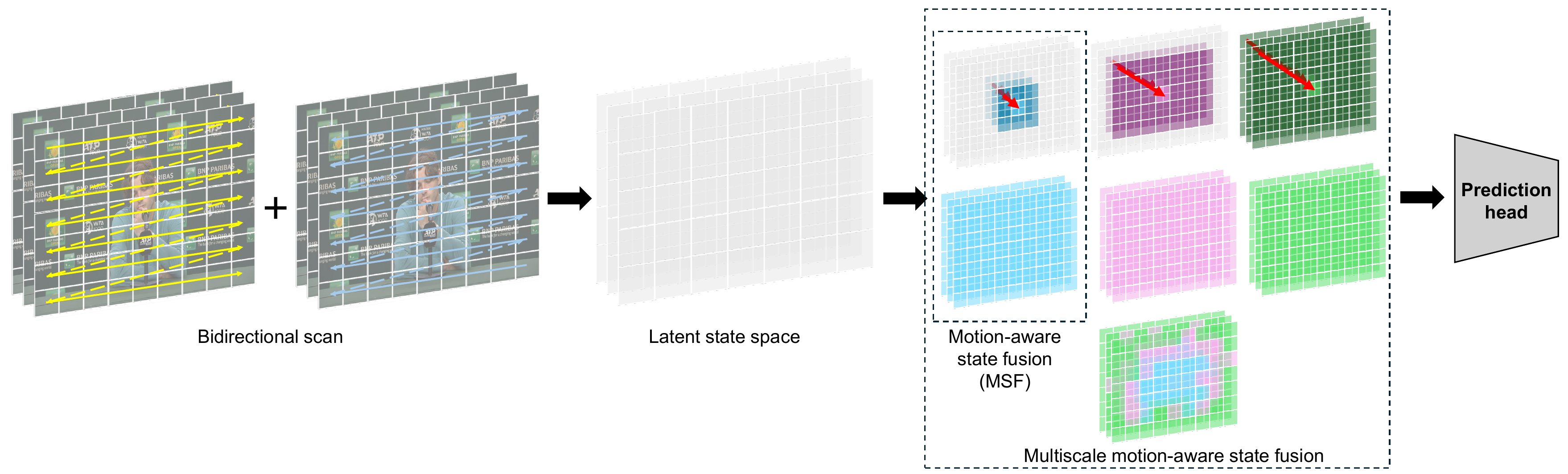}}\\
          \multicolumn{2}{c}{(c) The proposed MSF-Mamba.}

    \end{tabular}
    \caption{\textcolor{black}{Comparison between the previous Mamba-based visual models and the proposed MSF-Mamba in MGR. MSF-Mamba (c) enhances the original Mamba architecture by introducing multiscale motion-aware state fusion to aggregate local contextual states, and an adaptive scale weighting mechanism that dynamically weighs fused states at different scales. The red arrow denotes the interaction of state fusion.}}
    \label{fig:Motivation}
\end{figure*}

Recently, state space models (SSMs)~\cite{gu2021efficiently} have been explored as efficient alternatives to Transformer for visual tasks~\cite{Gong2025AVS,liu2024vmamba,huang2024localmamba,xiao2025frequency,pan2024mambasci,zhou2025binarized}. Mamba~\cite{gu2023mamba} is a selective SSM model that achieves linear-time processing while effectively capturing long-range dependencies. However, Mamba is essentially based on a recursive state-update mechanism\cite{qu2024survey}. The input of Mamba is a 1D sequence and can only process pixels (or patches) in scan order~\cite{xu2024visual}, which makes it insensitive to the spatiotemporal local structure in the video~\cite{huang2024localmamba}. As illustrated in Figure~\ref{fig:Motivation}, most \textcolor{black}{visual Mamba-based models}~\cite{liu2024vmamba,huang2024localmamba} attempt to address the lack of local spatiotemporal awareness by introducing different scanning strategies (e.g., selective scan or local scans), but these only reshape the input sequence order without fundamentally resolving this problem. Since the core Mamba architecture operates on 1D sequential modeling, these scans still fail to explicitly encode local spatiotemporal pixel (or patches) dependencies in the latent state space.

To address these limitations, we propose \textcolor{black}{motion-aware state fusion mamba (MSF-Mamba)}, a novel architecture that integrates motion-aware state fusion into the Mamba framework. Our design enables both local and global spatiotemporal modeling while maintaining the linear computational complexity of Mamba. Furthermore, \textcolor{black}{an multiscale version called MSF-Mamba$^{+}$ is proposed}. More precisely, we introduce a multiscale central frame difference state fusion module (MCFM), which injects local spatiotemporal awareness into the latent state space. In addition, we introduce an adaptive scale weighting module (ASWM), enabling the model to focus on the most relevant motion representations across different scales. We evaluate MSF-Mamba and its enhanced multiscale variant MSF-Mamba$^{+}$ \textcolor{black}{on two public datasets, namely, iMiGUE~\cite{liu2021imigue} and SMG~\cite{chen2023smg}.} The experimental results show that MSF-Mamba and MSF-Mamba$^{+}$ achieve state-of-the-art (SoTA) performance, outperforming existing CNN-, Transformer-, and SSM-based models while maintaining high efficiency. For example, MSF-Mamba$^{+}$ outperforms the baseline model VideoMamba~\cite{li2025videomamba} on SMG and iMiGUE by a large margin, \textcolor{black}{with Top-1 accuracy} by +2.9\% and +3.0\%, respectively. 

The contributions of this paper are summarized as follows: 1) We propose MSF-Mamba, a novel Mamba-based framework for MGR, which integrates motion-aware local spatiotemporal modeling into Mamba; 2) \textcolor{black}{We design MCFM to enhance Mamba by explicitly modeling local spatiotemporal patterns and introducing motion-awareness through multiscale central frame difference (CFD) operations}; 3) We develop ASWM to dynamically weigh motion-aware fused states, enabling the model to adaptively emphasize MG-relevant information at different scales; 4) We demonstrate that the proposed MSF-Mamba and MSF-Mamba$^{+}$ achieve SoTA performance on two public MGR datasets, namely, iMiGUE and SMG, showing superior performance and significantly high computational efficiency compared to CNN-, Transformer-, and SSM-based baseline models.

\section{Related work}
\subsection{Micro Gesture Recognition}

Early works \textcolor{black}{on} action recognition~\cite{li2025prototypical,guo2024mac} predominantly adopt CNN architectures. C3D~\cite{tran2015learning} \textcolor{black}{uses 3D convolutions to model spatiotemporal features in videos.} I3D~\cite{carreira2017quo} extends 3D convolutions with an inflated dual-stream design, combining RGB and optical flow streams to enhance motion representation. TSN~\cite{wang2016temporal} introduces sparse frame sampling to improve computational efficiency, while TSM~\cite{lin2019tsm} enhances temporal modeling by integrating a time-shift mechanism into 2D CNNs. MA-Net~\cite{guo2024benchmarking} achieves promising performance by integrating squeeze-and-excitation and temporal shift modules into a ResNet-50 backbone. However, CNN-based models are limited in modeling global dependencies due to their reliance on local convolutional operations, leading to suboptimal performance~\cite{xiao2025frequency}. Transformer architectures have demonstrated superior performance in global dependencies modeling through self-attention mechanisms. 
TimeSformer~\cite{bertasius2021space} applies divided space-time attention to model global spatiotemporal dependencies. \textcolor{black}{Some methods enhance the Transformer's ability to capture local spatiotemporal dependencies.} Video Swin Transformer (VSwin)~\cite{liu2022video} adopts a hierarchical sliding window approach to improve efficiency. UniformerV2~\cite{li2023uniformerv2} combines convolutional operations for local feature extraction with cross-attention modules for global context aggregation. However, these Transformer-based models typically incur quadratic computational complexity with respect to sequence length, leading to significant computational cost \textcolor{black}{and slow inference}~\cite{xiao2025frequency}. 

\subsection{State Space Models}

SSMs~\cite{gu2021efficiently} have recently emerged as a promising approach for modeling long-range dependencies with linear complexity. Mamba~\cite{gu2023mamba} enhances this framework by introducing \textcolor{black}{an input-conditioned parameterization mechanism} and selective scanning. \textcolor{black}{Building on the success of Mamba}, several works have extended Mamba to the vision domain~\cite{liu2024vmamba, xiao2025spatialmamba,huang2024localmamba}. For example, \textcolor{black}{VMamba~\cite{liu2024vmamba} applies Mamba to image classification through spatially structured scanning, while Local-Mamba~\cite{huang2024localmamba} strengthens locality via neighborhood-aware token updates. In the context of action recognition~\cite{li2025videomamba,2024videomambasuite}, VideoMamba~\cite{li2025videomamba} and Video-Mamba-Suite~\cite{2024videomambasuite} adapt Mamba by flattening spatiotemporal inputs into one-dimensional sequences, enabling efficient long-range spatiotemporal dependency modeling.} These methods enable efficient global modeling, but lack mechanisms for capturing local spatiotemporal dependencies. As shown in Table~\ref{tab:motivation}, CNN-based models are limited to local dependency modeling, while Transformer-based models suffer from quadratic complexity. SSM-based action recognition models provide efficient long-range modeling with linear complexity, yet lack the ability to model local dependencies and are not motion-aware. In contrast, our proposed MSF-Mamba simultaneously supports global and local spatiotemporal modeling, incorporates motion-awareness, and retains linear complexity, making it suited for MGR tasks.

\begin{table}[t]
    \centering
    \scriptsize
    \setlength\tabcolsep{1pt}
    \caption{
    Comparison of representative action recognition models. “Primarily local/global” refers to the dominant inductive bias. $n$ denotes the sequence length (i.e., number of video frames or tokens).
    }
    \begin{tabular}{lcccc}
    \toprule
    \multirow{2}{*}{\textbf{Method}} & \textbf{Model} & \textbf{Spatiotemporal} & \textbf{Comput.} & \textbf{Motion} \\
    &\textbf{Type}&\textbf{Dependency}&\textbf{Complexity}&\textbf{aware?}\\
    \midrule
    C3D~\cite{tran2015learning} & CNN & Primarily local & $O(n)$ & \xmark \\
    I3D~\cite{carreira2017quo} & CNN& Primarily local & $O(n)$ & \xmark \\
    TSM~\cite{lin2019tsm} & CNN& Local with temporal aggregation& $O(n)$ & \cmark \\
    TSN~\cite{wang2016temporal} & CNN& Primarily Local& $O(n)$ & \xmark \\
    Swin-T~\cite{liu2021swin} & Transformer & Local with global aggregation & $O(n \log n)$ & \xmark \\
    Uniformer~\cite{li2023uniformerv2} & Transformer & Global + local & $O(n^2)$ & \xmark \\
    VideoMamba~\cite{li2025videomamba} & SSM & Primarily global & $O(n)$ & \xmark \\
    MambaSuite~\cite{2024videomambasuite} & SSM &Primarily global& $O(n)$ & \xmark \\
    MSF-Mamba & SSM & Global + local & $O(n)$ & \cmark \\
    \bottomrule
    \end{tabular}
    \label{tab:motivation}
\end{table}
\section{Proposed Method}
\begin{figure*}
    \centering
    \includegraphics[width=0.9\linewidth]{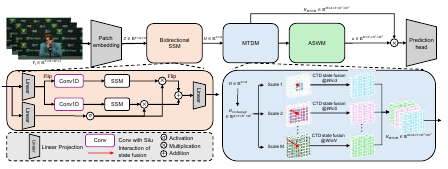}
     \caption{\textcolor{black}{Overview of the proposed MSF-Mamba framework for micro gesture recognition (MGR). Given an input video sequence \( V_i \in \mathbb{R}^{T \times H \times W \times 3} \), patch embedding generates token sequence \( Z \in \mathbb{R}^{n \times d} \). A Bidirectional SSM module to generate hidden states \( H \in \mathbb{R}^{n \times d} \). Multiscale central frame difference state fusion module (MCFM), which applies central temporal difference (CTD)-based state fusion over multiple window sizes to produce \( H_{\text{MCFM}} \in \mathbb{R}^{M \times d \times T \times H' \times W'} \). The adaptive scale weighting module (ASWM) then adaptively aggregates these multiscale fused states using attention weights $\alpha \in \mathbb{R}^{3 \times T \times H' \times W'}$. The final feature map is linearly projected for final prediction.}}
    \label{fig:framework}
\end{figure*}

\textcolor{black}{In this section, we discuss the proposed MSF-Mamba and its variant MSF-Mamba$^{+}$. For simplicity, both are collectively referred to as MSF-Mamba. The differences between the two will be presented in the experiments section.}

\subsection{Preliminaries}\label{sec:Preliminaries}

State space models (SSMs)~\cite{gu2021efficiently} are structured sequence models designed to capture long-range dependencies with linear computational complexity. They introduce hidden states $h(t) \in \mathbb{R}^N$ to model the evolution of input signals $x(t) \in \mathbb{R}^D$ over time and produce outputs $y(t) \in \mathbb{R}^{D'}$. In continuous time, a standard linear time-invariant SSM is formulated as follows:
\begin{align}
    \frac{d}{dt} h(t) &= A h(t) + B x(t), \\
    y(t) &= C h(t),
\end{align}
where $A \in \mathbb{R}^{N \times N}$, $B \in \mathbb{R}^{N \times D}$, and $C \in \mathbb{R}^{D' \times N}$ are learnable matrices governing the system dynamics.

To integrate SSMs into deep learning models, Mamba~\cite{gu2023mamba} discretizes this continuous system using the Zero-Order Hold (ZOH) method~\cite{pechlivanidou2022zero}, resulting in the following discrete transition matrices:
\begin{align}
    A_d &= \exp(\Delta A), \\
    B_d &= \Delta A^{-1} (A_d - I) B,
\end{align}
where $\Delta$ is the discretization step size and $I$ is the identity matrix. The system is then updated over discrete time steps as follows:
\begin{align}\label{eq:ssm}
    h_t &= A_d h_{t-1} + B_d x_t, \\
    y_t &= C h_t,
\end{align}
In Mamba~\cite{gu2023mamba}, the parameters $B_t$, $C_t$, and $\Delta_t$ are not fixed but are dynamically generated as input-dependent functions at each time step, computed as linear projections of the input as follows:
\begin{align}
    B_t,\, C_t,\, \Delta_t = f(x_t).
\end{align}

Unlike Transformers that rely on pairwise attention computation with $\mathcal{O}(n^2)$ computational complexity for sequence length $n$, Mamba leverages the recurrence-style update of SSMs. Each step only depends on the previous hidden state $h_{t-1}$ and current input $x_t$, leading to a per-step cost of $\mathcal{O}(1)$ and an overall complexity of $\mathcal{O}(n)$. This makes Mamba suitable for long sequence processing. While Mamba achieves efficient long-range sequence modeling, it lacks mechanisms to capture local spatiotemporal dependencies. To address this limitation, we propose MSF-Mamba, a novel framework that enhances Mamba with motion-aware state fusion for global and local spatiotemporal patterns modeling.

\subsection{Overview of MSF-Mamba}

As illustrated in Figure~\ref{fig:framework}, given an input video clip $V_i \in \mathbb{R}^{3 \times T \times H \times W}$, we first apply 3D patch embedding to extract spatiotemporal tokens $Z_p \in \mathbb{R}^{n \times d}$, where $n = T \cdot H' \cdot W'$, and $H'$, $W'$ denote the patch-wise spatial resolution. Then we add learnable positional embeddings, including temporal embeddings $P_t$ and spatial embeddings $P_s$, to retain positional information. The embedded sequence is formulated as follows:
\begin{equation}
    Z = Z_p + P_s + P_t.
\end{equation}
The embedded sequence $Z$ are then passed through a Bidirectional SSM module that generates latent states $H \in \mathbb{R}^{n \times d}$. To incorporate local spatiotemporal and motion-aware cues, the states $H \in \mathbb{R}^{n \times d}$ are fed into the multiscale central frame difference state fusion module (MCFM). \textcolor{black}{Each branch $k$ outputs a feature tensor $F^{(k)} \in \mathbb{R}^{d \times T \times H' \times W'}$, corresponding to a different scale.} The adaptive scale weighting module (ASWM) then adaptively aggregates these multiscale fused states using attention weights $\alpha \in \mathbb{R}^{3 \times T \times H' \times W'}$. The final feature map $F_{\text{final}} \in \mathbb{R}^{d \times T \times H' \times W'}$ is flattened and projected to produce the final representation for MG classification.

\subsection{Bidirectional SSM} 

\textcolor{black}{In the context of MGR, motion cues are subtle and often unfold dynamically along both forward and backward temporal directions. A unidirectional model processes information sequentially, which means it can only learn from past tokens. By contrast, bidirectional SSM can capture dependencies from both past and future tokens, allowing our model to form a more complete spatiotemporal understanding of the micro-gesture. Therefore, we adopt Bidirectional SSM for initial feature extraction.} Formally, given the input token sequence $Z \in \mathbb{R}^{n \times d}$, where $n = T \cdot H' \cdot W'$, we perform two independent SSMs to generate states: 1) Forward pass over the original sequence $Z$, yielding a hidden representation $H_{\rightarrow} \in \mathbb{R}^{n \times d}$; 2) Backward pass over the flipped version of the sequence, denoted as $\text{Flip}_T(Z)$, producing $H_{\leftarrow} \in \mathbb{R}^{n \times d}$. Each direction maintains its own set of SSM parameters, and both passes follow the discrete-time Mamba update rule as shown in Eq.~\eqref{eq:ssm}. The two directional outputs are then fused using simple averaging as
\begin{equation}   
H = \frac{1}{2} \left( H_{\rightarrow} + \text{Flip}_T(H_{\leftarrow}) \right),
\end{equation}
where $\text{Flip}_T(\cdot)$ reverses the sequence order of the backward output to align it with the forward sequence. Although this bidirectional SSM design improves global context aggregation over the flattened token sequence, it still lacks the capability to model localized motion patterns across space and time. Therefore, we introduce a motion-aware state fusion mechanism in the next subsection.

\subsection{Multiscale central frame difference state fusion module}

\textcolor{black}{To enhance motion awareness and incorporate Mamba’s local spatiotemporal modeling capability, we propose the multiscale central frame difference state fusion module (MCFM). In MCFM, we first reshape the latent states $H \in \mathbb{R}^{n \times d}$ from the bidirectional SSM into a structured 4D representation $F \in \mathbb{R}^{d \times T \times H' \times W'}$.} To enhance motion-awareness, we introduce a central frame difference (CFD) that explicitly captures temporal variation between adjacent frames of $F$. \textcolor{black}{Formally, at each timestamp $t$, we compute}
\begin{equation}
    D_t = F_t - \frac{1}{2}(F_{t-1} + F_{t+1}),
\end{equation}
where $D_t$ represents the latent states capturing the central frame difference. For the state fusion, we apply multiscale state fusion operators $\mathcal{S}_k(\cdot)$ to jointly model spatiotemporal representation at different window sizes. Let $X \in \mathbb{R}^{d \times T \times H' \times W'}$ denote either the original representation $F$ or the motion-aware representation $D$. Each operator is defined as follows:
\begin{equation}
    \mathcal{S}_k(X)_t = \sum_{(\delta\tau, \delta h, \delta w) \in \mathcal{N}_k} W_k(\delta\tau, \delta h, \delta w) \cdot X_{t+\delta\tau, h+\delta h, w+\delta w},
\end{equation}
where $\mathcal{N}_k \subset [-r_k, r_k]^3$ defines a local 3D neighborhood for scale (window size) $k$ (e.g., $r_k = 3, 5, 7$). The ablation study of the selection of window size is presented in Table~\ref{tab:windowsizeselection}. $W_k \in \mathbb{R}^{(2r_k+1)^3}$ is a learnable kernel weight tensor. The operation performs local weighted aggregation over a cube centered at each location $(t,h,w)$. Each fusion output at scale $k$ is computed by combining the original and motion-aware representation:
\begin{equation}
    F^{(k)} = \mathcal{S}_k(F) + \theta_k \cdot \mathcal{S}_k(D),
\end{equation}
where $\theta_k \in (0,1)$ is a learnable scalar gate that modulates the contribution of the motion signal. The initial value of $\theta_k$ is 0.5. \textcolor{black}{The value of 0.5 was determined through hyperparameter tuning using PyTorch.} The outputs $\{F^{(win@3\times3\times3)}, F^{(win@5\times5\times5)}, F^{(win@7\times7\times7)}\}$ form a set of multiscale motion-aware fused states enriched with local motion dynamics. The ablation study of the proposed MCFM and CFD is presented in Table~\ref{tab:ablation_scan}. As shown in Figure~\ref{fig:CFD}, we compare the average activation maps of the output layer with and without CFD in MCFM, highlighting the contribution of motion-aware modeling.

\subsection{Adaptive Scale Weighting
Module}\label{sec:ASWM}

To adaptively integrate \textcolor{black}{the fused multiscale motion-aware states obtained from MCFM}, we introduce the adaptive scale weighting
module (ASWM). \textcolor{black}{Whereas naive average fusion treats all scales equally, ASWM dynamically assigns spatiotemporal attention weights, enabling the model to emphasize the most relevant scale at each position.} Given the multiscale representation $\{F^{(win@3\times3\times3)}, F^{(win@5\times5\times5)}, F^{(win@7\times7\times7)}\}$, where each $F^{(k)} \in \mathbb{R}^{d \times T \times H' \times W'}$, we first concatenate them along the channel dimension to form a unified representation $F_{\text{concat}} = \text{Concat}(F^{(win@3\times3\times3)}, F^{(win@5\times5\times5)}, F^{(win@7\times7\times7)}) \in \mathbb{R}^{3d \times T \times H' \times W'}$. We then use a two-layer 3D convolutional sub-network to compute soft attention weights for each scale as follows:

\begin{equation}
A = \text{Conv3D}_{\text{attn}}(F_{\text{concat}}) \in \mathbb{R}^{3 \times T \times H' \times W'},
\end{equation}
where $\text{Conv3D}_{\text{attn}}$ consists of two stacked convolutional layers with ReLU and softmax. The output $A$ represents the attention logits over the three scales at each spatiotemporal position and is normalized across the scale dimension as

\begin{equation}
\alpha_{k,t,h,w} = \frac{\exp(A_{k,t,h,w})}{\sum_{k'=1}^3 \exp(A_{k',t,h,w})}.
\end{equation}
Let $\alpha \in \mathbb{R}^{3 \times T \times H' \times W'}$ be the soft attention weight map, and stack the input features into 

\begin{equation}
F_{\text{stack}} \in \mathbb{R}^{3 \times d \times T \times H' \times W'}.
\end{equation}
The final feature is computed as a weighted sum across the scale dimension as follows:
\begin{equation}
F_{\text{final}}(c, t, h, w) = \sum_{k=1}^3 \alpha_{k,t,h,w} \cdot F_{\text{stack}}(k, c, t, h, w),
\end{equation}
The output $F_{\text{final}} \in \mathbb{R}^{d \times T \times H' \times W'}$ serves as the final motion-aware representation and is passed directly to the classification head after flattening. The pseudocode of ASWM is present in Algorithm~\ref{alg:aswm}.

\begin{algorithm}
\footnotesize
\caption{\textcolor{black}{Adaptive Scale Weighting Module (ASWM)}}
\label{alg:aswm}
\begin{algorithmic}[1]
\Require Multiscale feature tensors $\{F^{(k)}\}_{k=1}^3$, where $F^{(k)} \in \mathbb{R}^{d \times T \times H' \times W'}$ corresponds to window size $k \in \{3, 5, 7\}$.
\Ensure Final weighted feature map $F_{\text{final}} \in \mathbb{R}^{d \times T \times H' \times W'}$.

\Statex \Comment{\textbf{Step 1: Concatenate multiscale features}}
\State $F_{\text{concat}} \gets \text{Concatenate}(F^{(3)}, F^{(5)}, F^{(7)})$ \Comment{Shape: $\mathbb{R}^{3d \times T \times H' \times W'}$}

\Statex \Comment{\textbf{Step 2: Compute attention weights}}
\State $A \gets \text{Conv3D}_{\text{attn}}(F_{\text{concat}})$ \Comment{Shape: $\mathbb{R}^{3 \times T \times H' \times W'}$}
\State $\alpha \gets \text{Softmax}(A)$ \Comment{Normalize across scale dimension}

\Statex \Comment{\textbf{Step 3: Stack features for weighted fusion}}
\State $F_{\text{stack}} \gets \text{Stack}(F^{(3)}, F^{(5)}, F^{(7)})$ \Comment{Shape: $\mathbb{R}^{3 \times d \times T \times H' \times W'}$}

\Statex \Comment{\textbf{Step 4: Apply attention and sum}}
\State $F_{\text{weighted}} \gets \text{Broadcast}(\alpha) \times F_{\text{stack}}$
\State $F_{\text{final}} \gets \text{Sum}(F_{\text{weighted}})$ \Comment{Sum across the scale dimension}

\State \Return $F_{\text{final}}$
\end{algorithmic}
\end{algorithm}

\subsection{Classification Head \& Loss}

Following ASWM, the output representation \textcolor{black}{is first flattened along the spatiotemporal dimensions.} To aggregate information from all locations, we apply global average pooling over the combined sequence, resulting in a feature vector. This vector is then passed through a linear classifier to produce the final prediction $\hat{y}_j$. The entire model is trained in an end-to-end fashion using the standard cross-entropy loss between the predicted class probabilities and the ground-truth gesture labels. Thus, the cross-entropy loss function is computed as
\begin{equation}
    \mathcal{L} = - \sum_{c=1}^{C} q_c \cdot \log \left( \frac{\exp(\hat{y}_c)}{\sum_{j=1}^{C} \exp(\hat{y}_j)} \right),
\end{equation}
where $\hat{y}_c$ is the predicted logit for class $c$. \textcolor{black}{$q_c$ denotes the one-hot ground-truth label distribution, with $q_c = 1$ for the correct class and $q_c = 0$ otherwise.}

\begin{figure*}
    \centering
    \tiny
    \setlength\tabcolsep{-1.5pt}
    \begin{tabular}{ccccccccc}
         \includegraphics[height=1.7cm,width=0.11\linewidth]{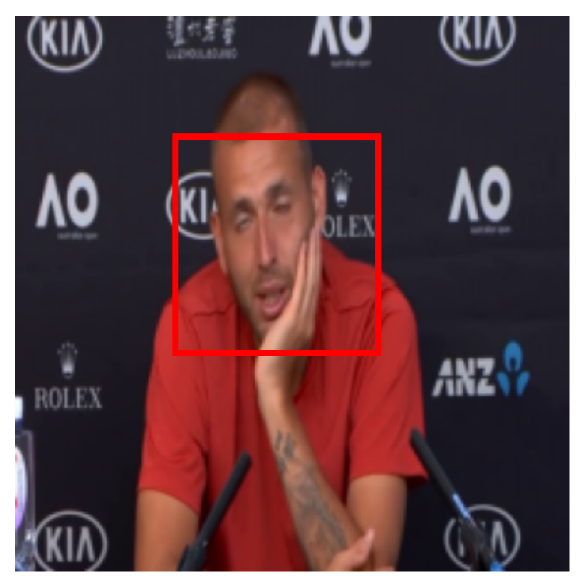}&\includegraphics[height=1.7cm,width=0.11\linewidth]{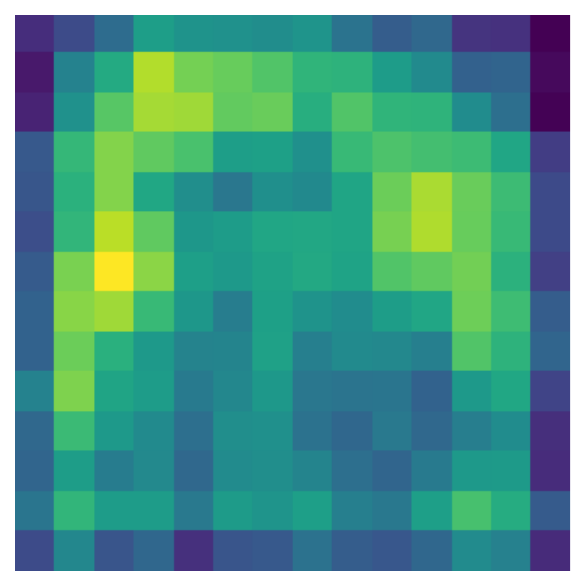}&\includegraphics[height=1.7cm,width=0.11\linewidth]{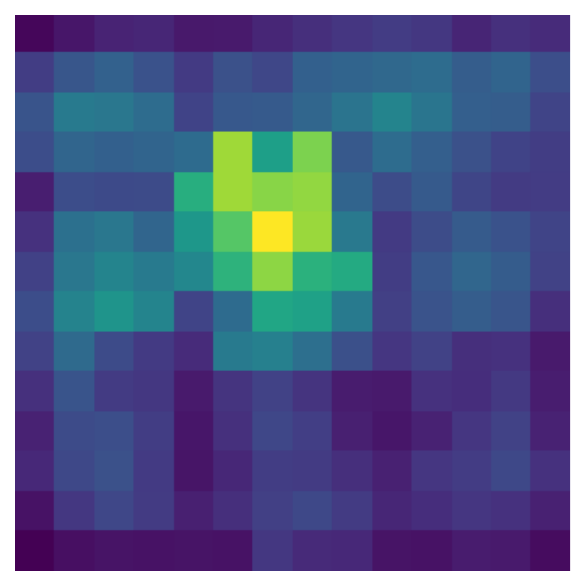}&\includegraphics[height=1.7cm,width=0.11\linewidth]{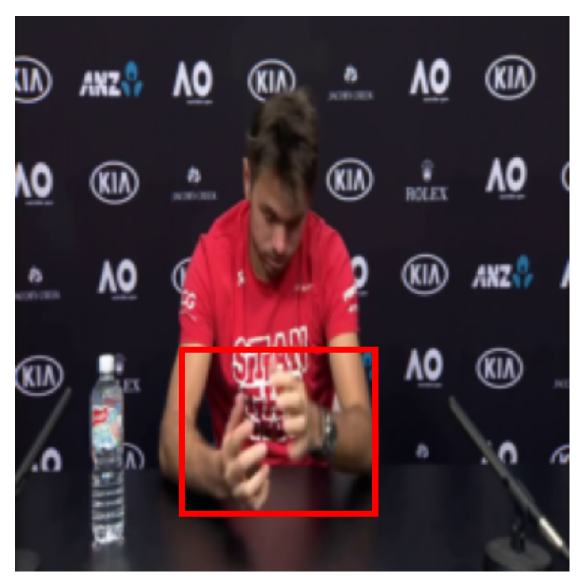}&\includegraphics[height=1.7cm,width=0.11\linewidth]{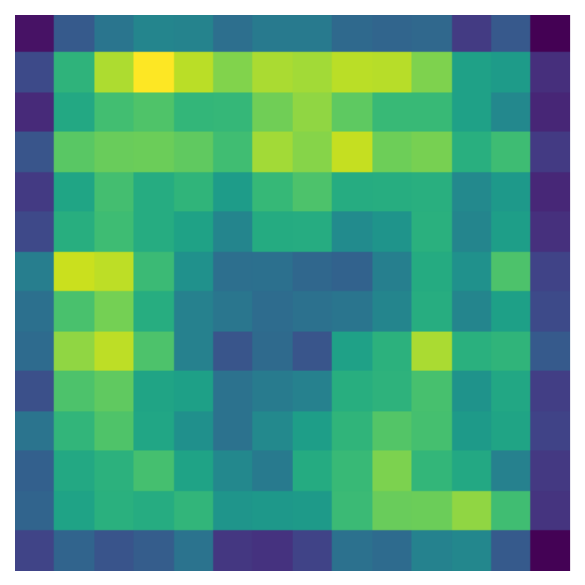}&\includegraphics[height=1.7cm,width=0.11\linewidth]{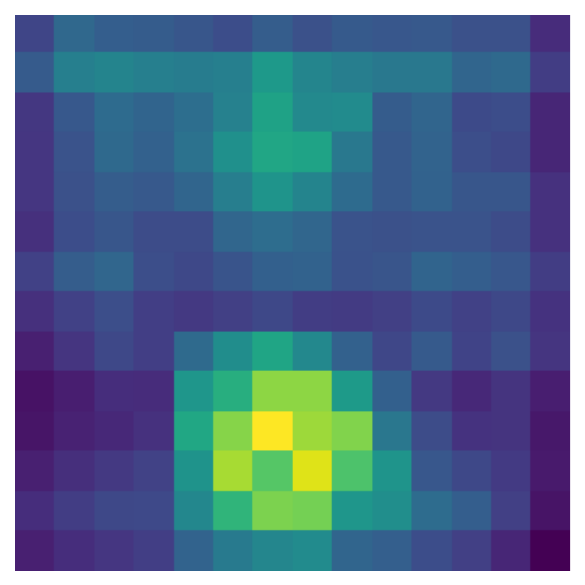}&\includegraphics[height=1.7cm,width=0.11\linewidth]{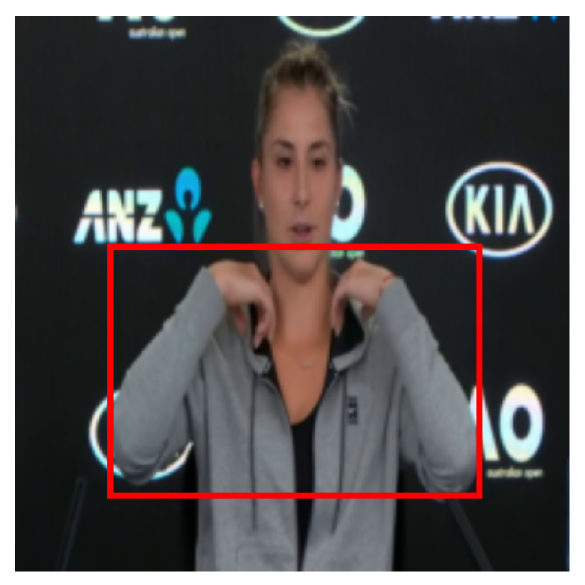}&\includegraphics[height=1.7cm,width=0.11\linewidth]{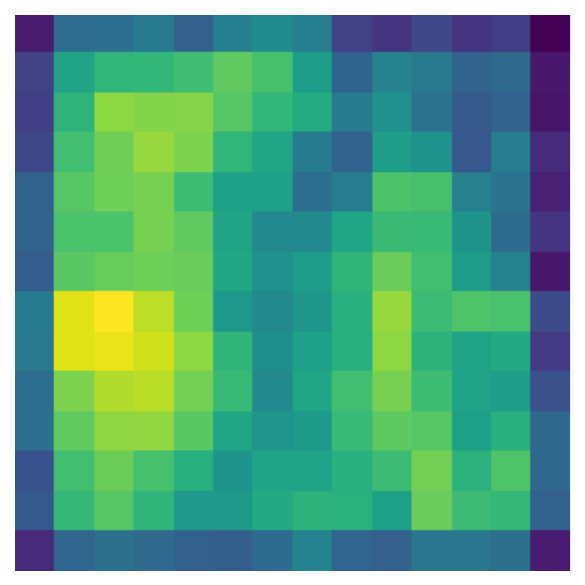}&\includegraphics[height=1.7cm,width=0.11\linewidth]{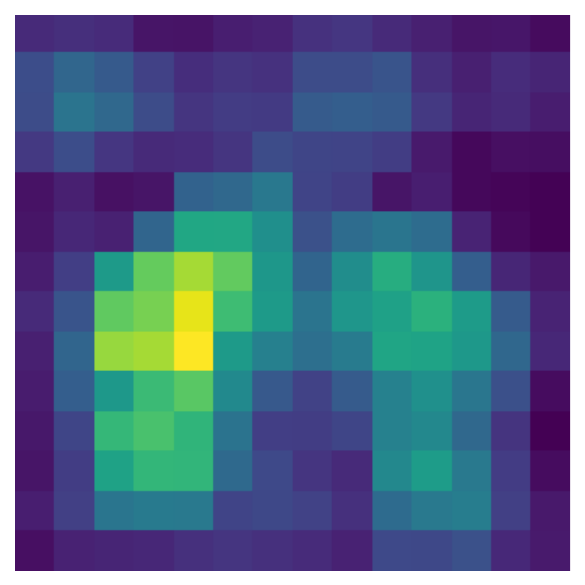}  \\
         (a) Touching facial parts&MCFM w/o CFD& MCFM w/ CFD&(b) Crossing fingers&MCFM w/o CFD& MCFM w/ CFD&(c) Pulling shirt collar&MCFM w/o CFD& MCFM w/ CFD\\
         \includegraphics[height=1.7cm,width=0.11\linewidth]{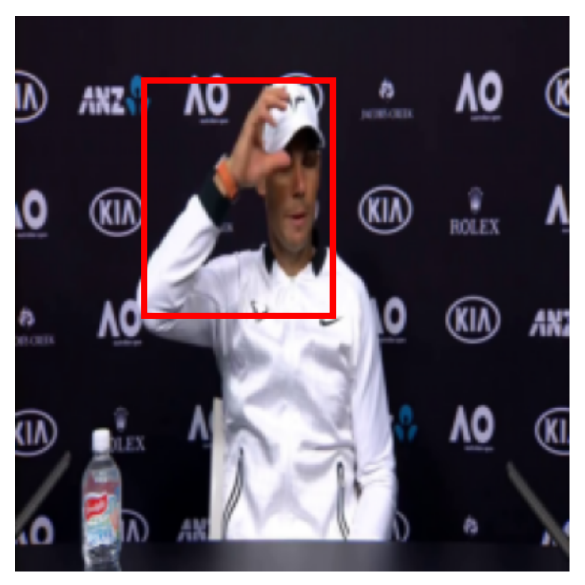}&\includegraphics[height=1.7cm,width=0.11\linewidth]{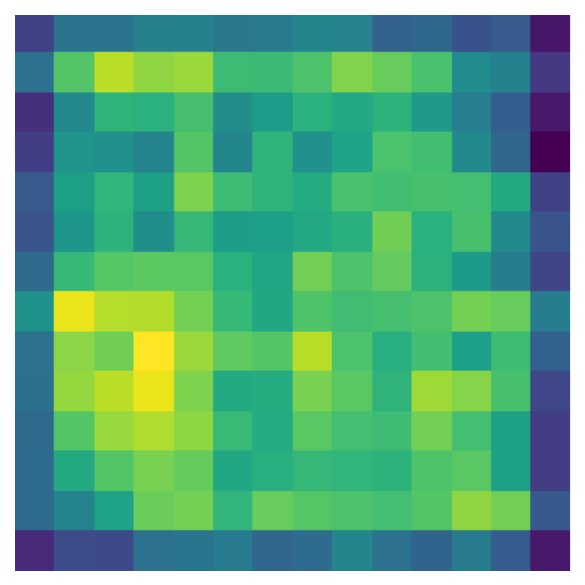}&\includegraphics[height=1.7cm,width=0.11\linewidth]{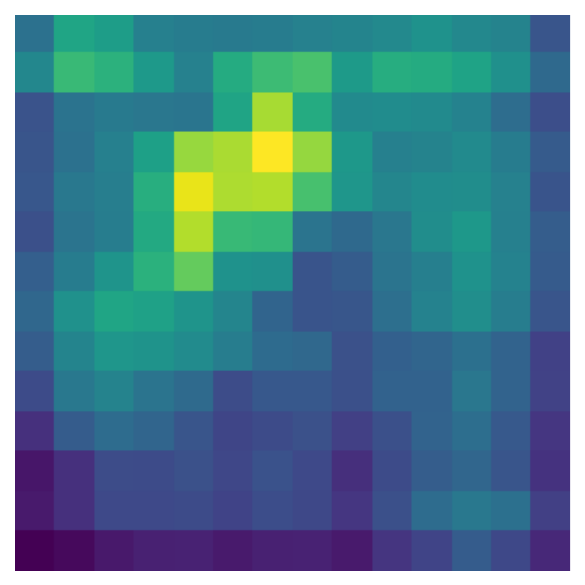}&\includegraphics[height=1.7cm,width=0.11\linewidth]{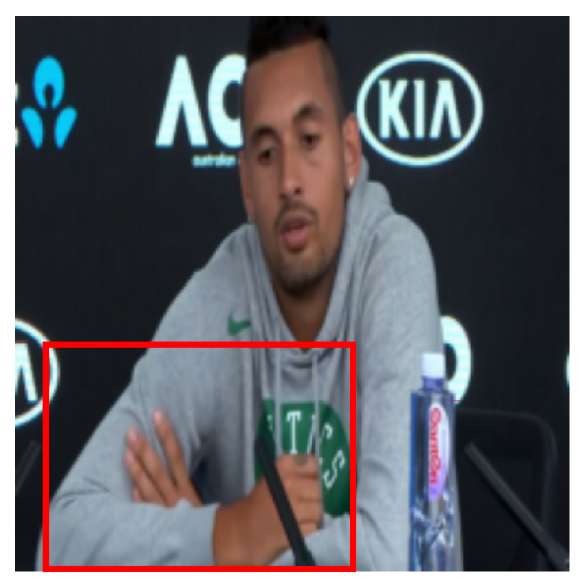}&\includegraphics[height=1.7cm,width=0.11\linewidth]{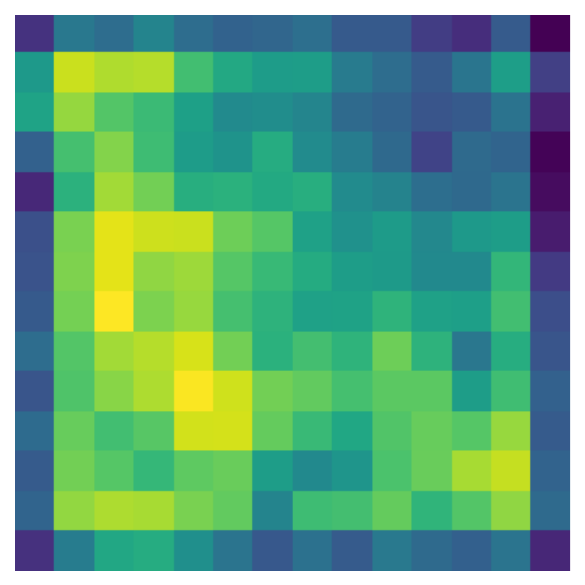}&\includegraphics[height=1.7cm,width=0.11\linewidth]{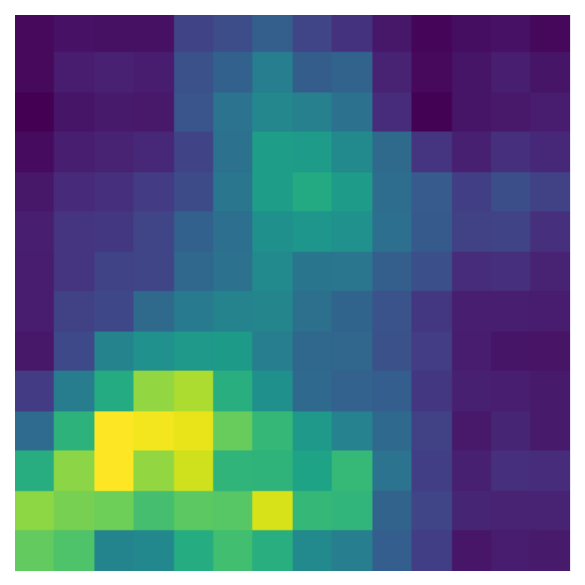}&\includegraphics[height=1.7cm,width=0.11\linewidth]{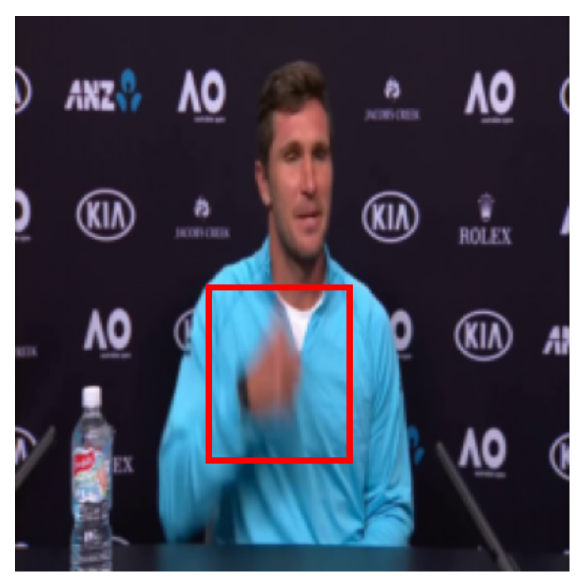}&\includegraphics[height=1.7cm,width=0.11\linewidth]{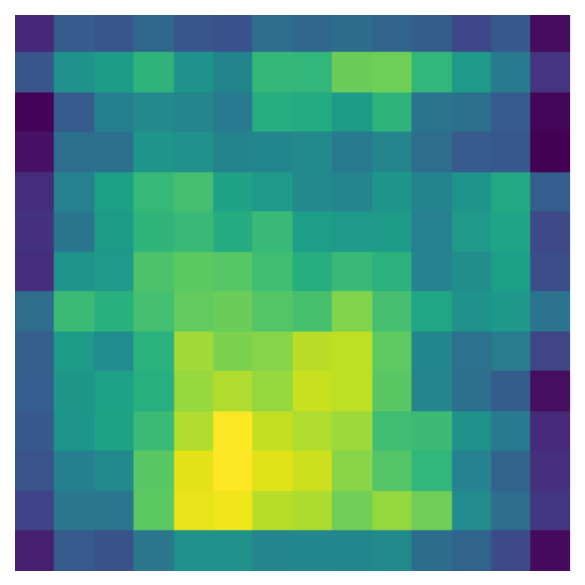}&\includegraphics[height=1.7cm,width=0.11\linewidth]{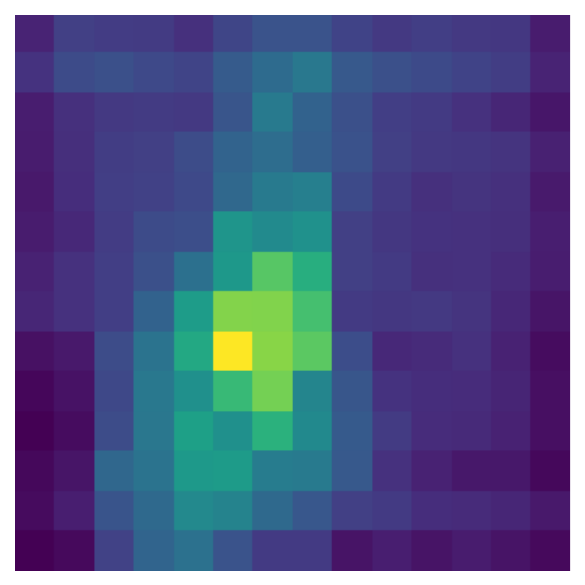}  \\
         (d) Touching hat&MCFM w/o CFD& MCFM w/ CFD&(e) Moving torso&MCFM w/o CFD& MCFM w/ CFD&(f) Illustrative gestures&MCFM w/o CFD& MCFM w/ CFD\\
         \includegraphics[height=1.7cm,width=0.11\linewidth]{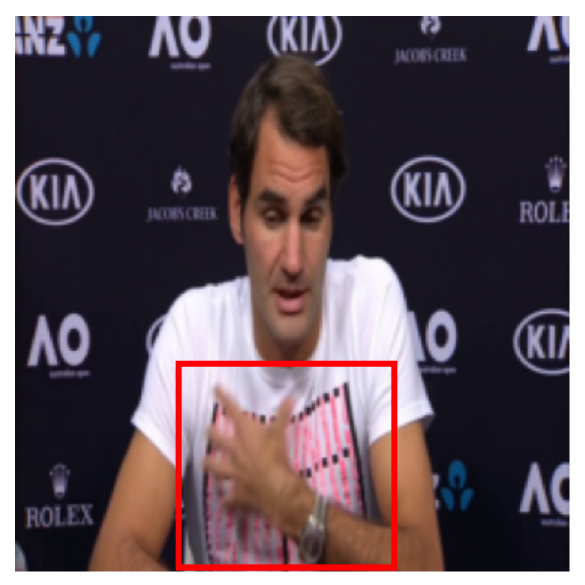}&\includegraphics[height=1.7cm,width=0.11\linewidth]{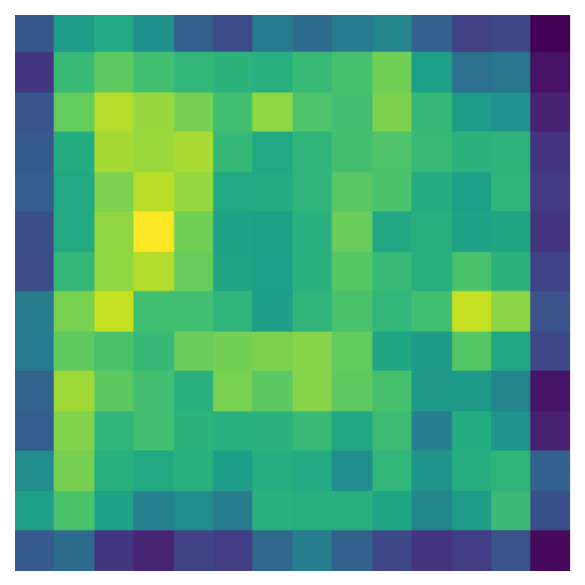}&\includegraphics[height=1.7cm,width=0.11\linewidth]{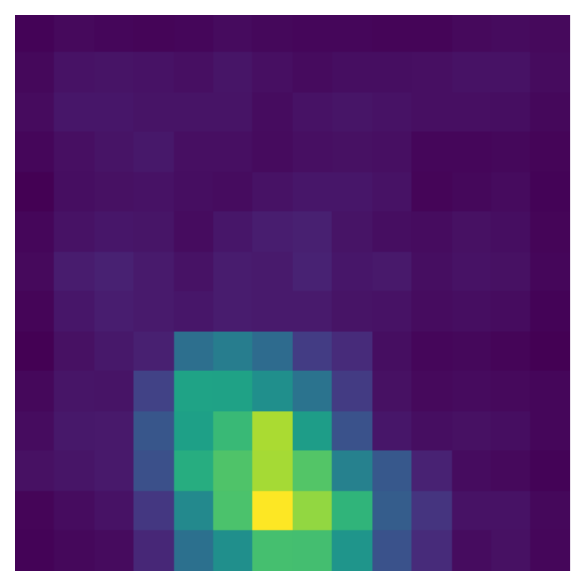}&\includegraphics[height=1.7cm,width=0.11\linewidth]{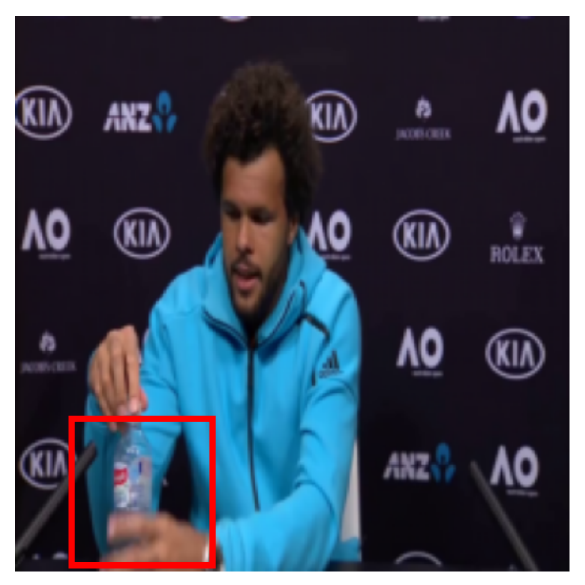}&\includegraphics[height=1.7cm,width=0.11\linewidth]{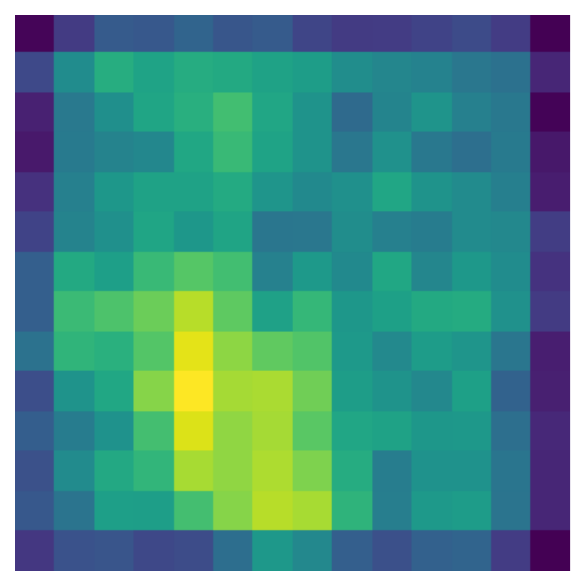}&\includegraphics[height=1.7cm,width=0.11\linewidth]{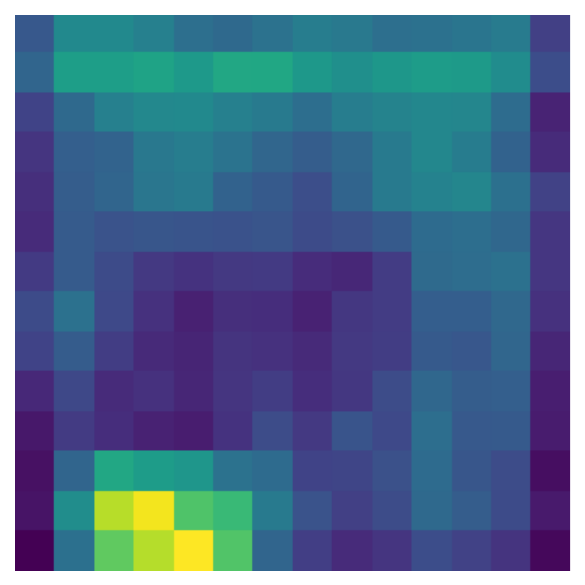}&\includegraphics[height=1.7cm,width=0.11\linewidth]{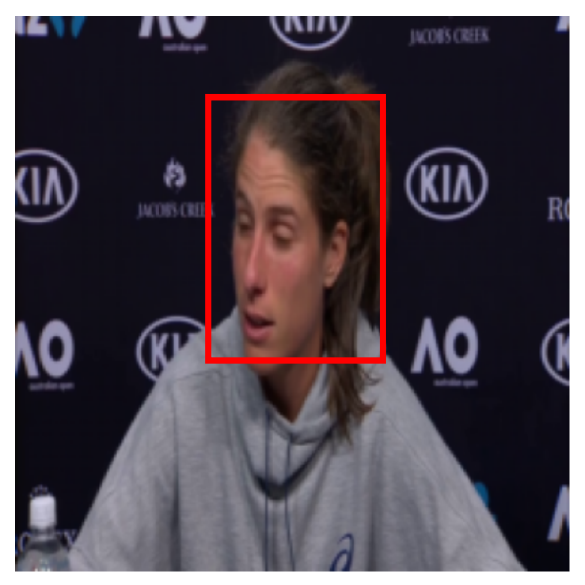}&\includegraphics[height=1.7cm,width=0.11\linewidth]{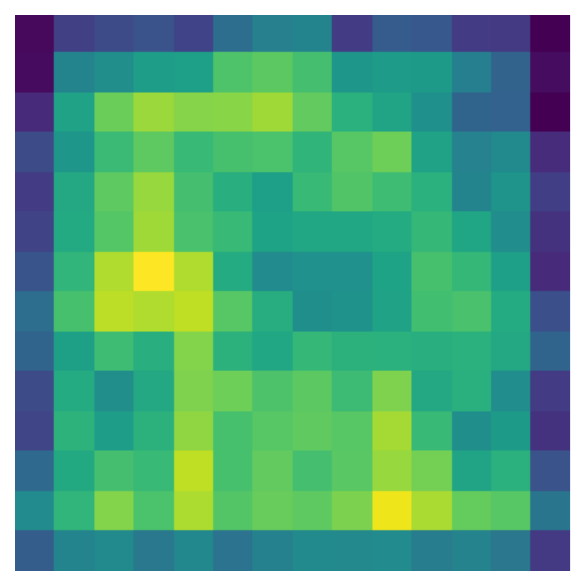}&\includegraphics[height=1.7cm,width=0.11\linewidth]{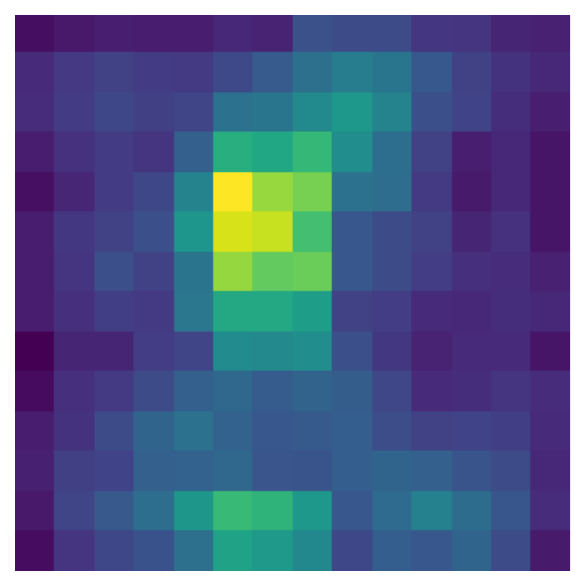}  \\
         (g) Supernatural notch&MCFM w/o CFD& MCFM w/ CFD&(h) Manipulating objects&MCFM w/o CFD& MCFM w/ CFD&(i) Head up&MCFM w/o CFD& MCFM w/ CFD\\
         \includegraphics[height=1.7cm,width=0.11\linewidth]{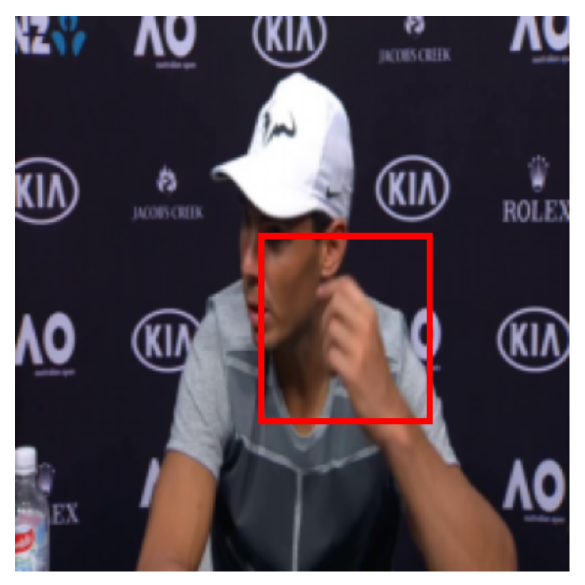}&\includegraphics[height=1.7cm,width=0.11\linewidth]{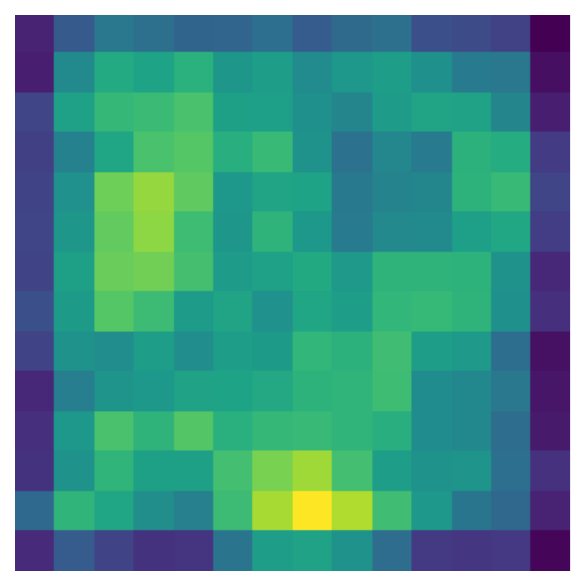}&\includegraphics[height=1.7cm,width=0.11\linewidth]{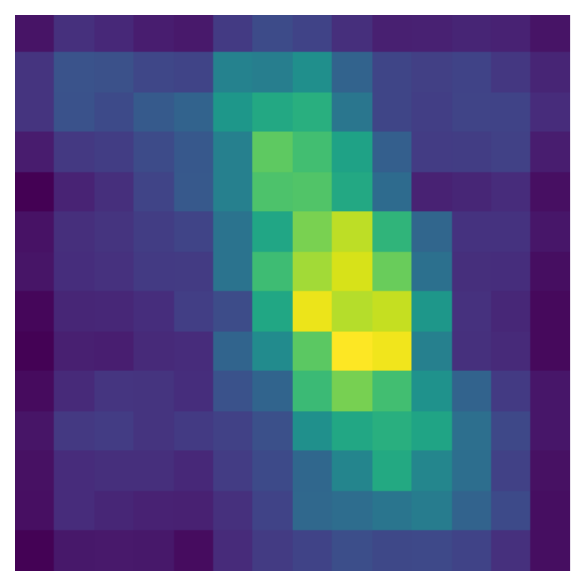}&\includegraphics[height=1.7cm,width=0.11\linewidth]{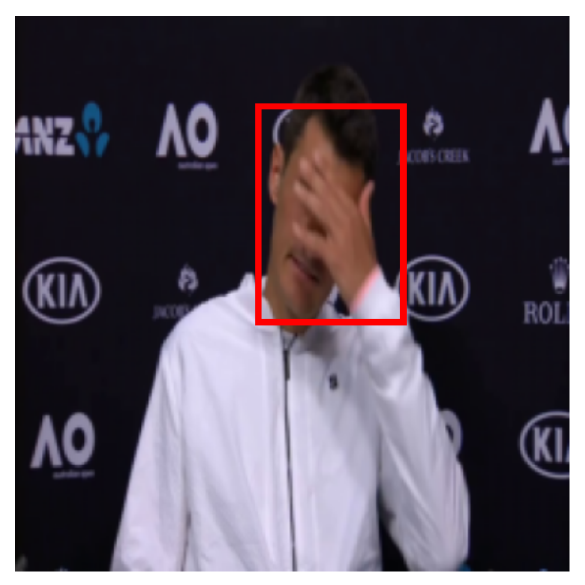}&\includegraphics[height=1.7cm,width=0.11\linewidth]{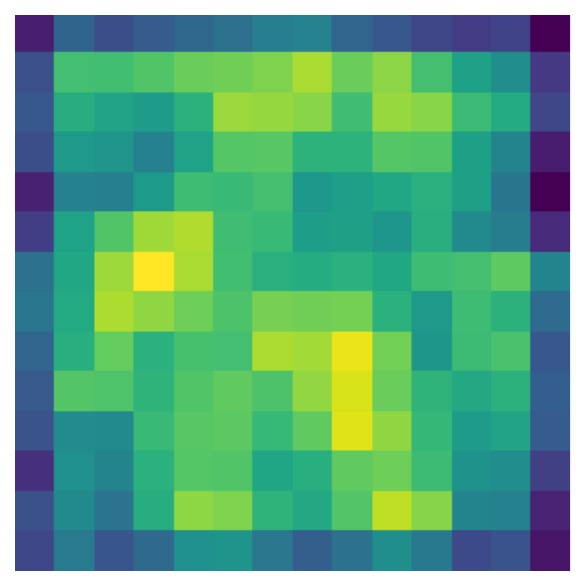}&\includegraphics[height=1.7cm,width=0.11\linewidth]{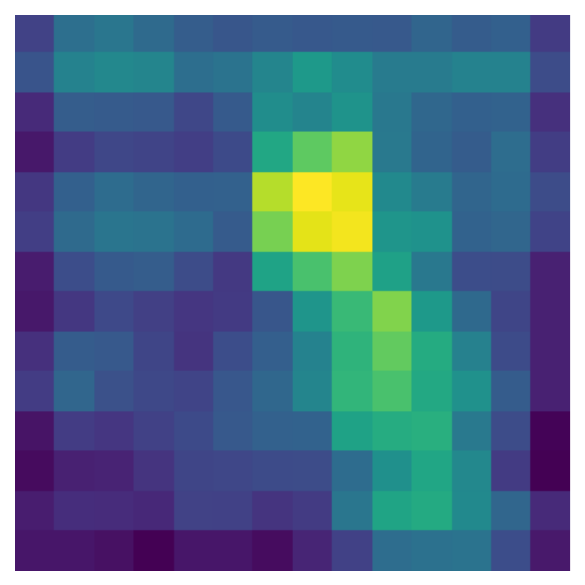}&\includegraphics[height=1.7cm,width=0.11\linewidth]{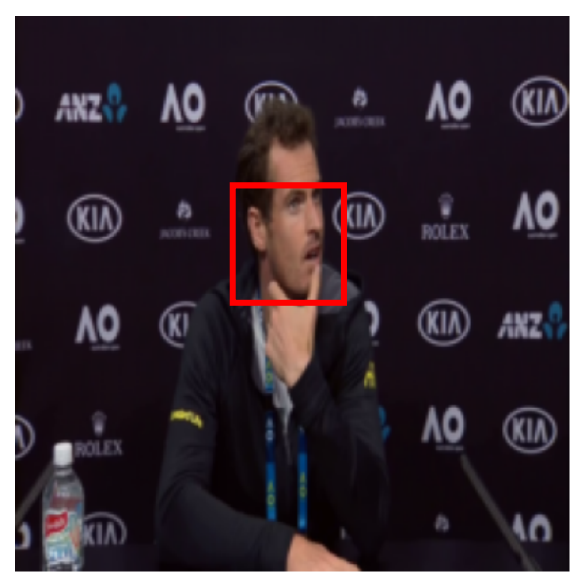}&\includegraphics[height=1.7cm,width=0.11\linewidth]{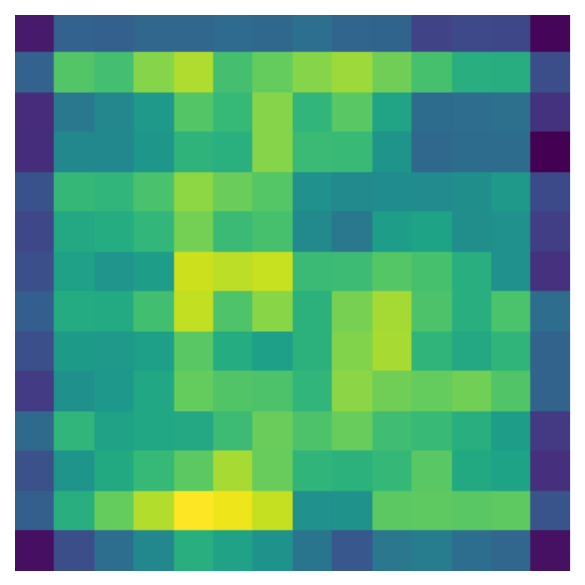}&\includegraphics[height=1.7cm,width=0.11\linewidth]{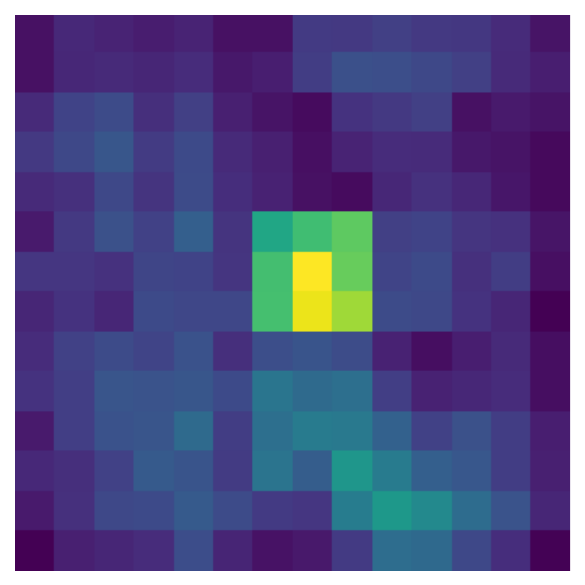}  \\
         (j) Touching ears&MCFM w/o CFD& MCFM w/ CFD&(k) Touching head&MCFM w/o CFD& MCFM w/ CFD&(l) Touching jaw&MCFM w/o CFD& MCFM w/ CFD
    \end{tabular}
    \caption{Qualitative visualization of activation map from the output of the multiscale central frame difference state fusion module (MCFM) across various micro gesture categories. Each subfigure group shows the raw input frame (left), the activation map of MCFM without central frame difference (CFD) (middle), and the activation map of MCFM with CFD (right).}
    \label{fig:CFD}
\end{figure*}

\section{Experiments}
\subsection{Experimental Settings}
\textbf{Datasets.} We evaluate our method on two public MGR datasets, namely, iMiGUE~\cite{liu2021imigue} and SMG~\cite{chen2023smg}, each featuring diverse MG types and collection settings. iMiGUE is a web-sourced dataset with 18,499 samples covering 32 MG categories. Videos are captured at a resolution of 1280×720 and have an average duration of 2.6 seconds. SMG is a lab-collected dataset containing 3,712 samples across 17 MG categories, with high-resolution video at 1920×1080 and an average duration of 1.8 seconds.

\textbf{Training.} We follow the training setup of Video-Mamba~\cite{li2025videomamba} to initialize our models with ImageNet-1K~\cite{deng2009imagenet} pre-training. \textcolor{black}{A 5-epoch warm-up is applied to stabilize early training, followed by 30 epochs in total to ensure convergence.} L2 regularization with a weight decay of 0.05 is applied to control model complexity. \textcolor{black}{The learning rate is linearly scaled with the batch size according to} \(4 \times 10^{-4} \times \frac{\text{batch size}}{256}\), and the Adamw optimizer ~\cite{loshchilov2018decoupled} is used for optimization. All experiments are conducted on a single NVIDIA A100 GPU. 

\textbf{Metrics.} Following the evaluation protocols of iMiGUE~\cite{liu2021imigue} and SMG~\cite{chen2023smg}, we report Top-1 and Top-5 accuracy as the evaluation metrics for MGR: 
\begin{equation}
\text{Top-}k\ \text{Accuracy} = \frac{1}{N} \sum_{i=1}^{N} \delta(y_i \in \hat{Y}_i^{(k)})
\end{equation}
where \(N\) is the number of samples, \(y_i\) is the ground-truth label of the \(i\)-th sample, and \(\hat{Y}_i^{(k)}\) denotes the top-\(k\) predicted labels. The function \(\delta(\cdot)\) returns 1 if the condition is true (i.e., the correct label is within the top-\(k\) predictions).

\textbf{Comparative methods.} To thoroughly evaluate the effectiveness of our proposed MSF-Mamba, we compare it against a range of SoTA RGB-based action recognition models. Specifically, we include CNN-based models such as C3D~\cite{tran2015learning}, I3D~\cite{carreira2017quo}, TSN~\cite{wang2016temporal}, TSM~\cite{lin2019tsm} and MA-Net~\cite{guo2024benchmarking}; Transformer-based models including TimeSformer~\cite{bertasius2021space}, Video Swin Transformer (VSwin)\cite{liu2022video}, and UniformerV2\cite{li2023uniformerv2}; and an SSM-based model, VideoMamba~\cite{li2025videomamba}. To ensure a fair comparison, all baseline methods are implemented and evaluated using the MMAction2~\footnote{https://github.com/open-mmlab/mmaction2} framework. \textcolor{black}{Specifically, we use a uniform sampling strategy for all models. For training and validation, we sample a single clip of 16 frames from each video. The training data pipeline includes random resized cropping (to a final size of $224\times224$) and random horizontal flipping with a probability of 0.5.} Other recent SSM models, such as VMamba~\cite{liu2024vmamba} and LocalMamba~\cite{huang2024localmamba} are specifically designed for image understanding tasks and thus are not included in our comparison experiment. Nevertheless, we adopt their proposed scanning strategies in our ablation studies as shown in Table~\ref{tab:ablation_scan}.

\subsection{Ablation Study}

To better understand the effectiveness of each component in MSF-Mamba, we conduct a series of ablation experiments on the SMG datasets. 

\textbf{Is motion-aware state fusion effective?} We evaluate the impact of our proposed MCFM and CFD on the SMG dataset under various scan strategies. This setting of MCFM and CFD only uses a single state fusion branch ($win@3\times3\times3$), without multiscale enhancement. For fair comparison, we also implement and test different scan orders, including bidirectional, local~\cite{huang2024localmamba}, and selective~\cite{liu2024vmamba} scanning. As shown in Table~\ref{tab:ablation_scan}, different scan strategies (e.g., bidirectional, local, and selective) yield similar Top-1 accuracies (53.18\%–53.54\%), indicating that the choice of scan order alone has limited influence on MGR performance. However, once the motion-aware fusion components are introduced, performance increases become more evident. Adding MCFM to the bidirectional scan leads to a clear gain, pushing Top-1 accuracy to 54.04\%. When combined with CFD, which explicitly emphasizes dynamic motion cues, our model achieves a Top-1 accuracy of 54.73\%, outperforming all other configurations. These results confirm that motion-aware state fusion is the key factor in enhancing the recognition performance of the Mamba-based model for fine-grained MGs. As shown in Figure~\ref{fig:CFD}, the qualitative comparison shows that incorporating CFD helps the model focus more on dynamic regions. \textcolor{black}{For example, "\textit{Touching facial parts}" and "\textit{Touching hat}" (Figure~\ref{fig:CFD} (a) and (d)) with CFD, the attention becomes highly localized and concentrated on the specific regions of the face or hat where the subtle touching motion occurs. This validates that CFD is crucial for capturing the motion of MGs. In addition, we conduct an ablation study comparing the CFD of MCFW with a first-order difference. As shown in Table~\ref{tab:Motion}, CFD outperforms the first-order difference. This demonstrates that CFD is more effective at modeling the subtle dynamics characteristic of MGs.}

\begin{table}[t]
\centering
\setlength\tabcolsep{1pt}
\caption{Ablation study on different scan strategies and the effect of motion-aware fusion on the SMG dataset.} 
\label{tab:ablation_scan}
\footnotesize
\begin{tabular}{c|cc|cc}
\toprule
\textbf{Scan Type} & \textbf{MCFM}&\textbf{CFD} & \textbf{Acc. Top-1(\%)}$\uparrow$& \textbf{Acc. Top-5(\%)}$\uparrow$ \\
\midrule
 Local (LocalMamba~\cite{huang2024localmamba})&\xmark&\xmark&53.54&87.79\\
 Selective (VMamba~\cite{liu2024vmamba})& \xmark &\xmark&53.19&87.31  \\\midrule
 Bidirectional&\xmark&\xmark&53.18&87.54  \\
 Bidirectional & \cmark&\xmark &54.04&88.78 \\
 Bidirectional & \cmark&\cmark &\textbf{54.73}&\textbf{89.35} \\
\bottomrule
\end{tabular}
\end{table}

\begin{table}[t]
    \centering
    \setlength\tabcolsep{2pt}
    \footnotesize
    \caption{\textcolor{black}{Ablation study of different motion modeling of MCFM on the SMG dataset.}}
    \begin{tabular}{ccc}\toprule
         \textbf{Motion Modeling}&\textbf{Acc. Top-1(\%)}$\uparrow$& \textbf{Acc. Top-5(\%)}$\uparrow$ \\\midrule
         First-order Difference&54.39&89.11\\
         Central Frame Difference&\textbf{54.73}&\textbf{89.35}\\\bottomrule
    \end{tabular}
    \label{tab:Motion}
\end{table}

\textbf{Do multiscale representations help?} To evaluate the effectiveness of multiscale modeling, we conduct an ablation study \textcolor{black}{by varying the number of fused state branches used in MCFM} on the SMG dataset. The different scale branches are aggregated by average sum. As shown in Table~\ref{tab:windowsizeselection}, each branch corresponds to a specific window size ranging from $win@3\times3\times3$ to $win@9\times9\times9$. \textcolor{black}{We first evaluate single-scale models, where the $win@7\times7\times7$ configuration achieves the best Top-1 accuracy of 55.07\%.} In contrast, the largest $win@9\times9\times9$ windows perform slightly worse. Building on this, we explore multiscale motion-aware modeling. Adding multiple branches consistently improves performance, with the combination of $win@3\times3\times3$, $win@5\times5\times5$, and $win@7\times7\times7$ windows achieving the best Top-1 accuracy of 55.60\% and Top-5 of 89.50\%. Interestingly, introducing a fourth scale $win@9\times9\times9$ slightly degrades performance to 55.38\%. We hypothesize that the large window size \textcolor{black}{may incorporate irrelevant context (e.g., background), thereby introducing unnecessary noise.} The experimental results suggest that while multiscale modeling is beneficial, there exists an optimal range of window sizes for state fusion for MGR. In our case, the $win@3\times3\times3$, $win@5\times5\times5$, and $win@7\times7\times7$ configurations provide the best balance. 
\begin{table}[ht]
    \centering
    \footnotesize
    \setlength\tabcolsep{2pt}
    \caption{Ablation study on the different window size selection on multiscale MCFM on the SMG dataset.}
    \begin{tabular}{c|c|c|c|cc}
    \toprule
         \multicolumn{4}{c}{\textbf{Window sizes}}&\textbf{Accuracy}& \textbf{Accuracy} \\
         $3\times3\times3$& $5\times5\times5$& $7\times7\times7$& $9\times9\times9$&\textbf{Top-1(\%)$\uparrow$}&\textbf{Top-1(\%)$\uparrow$}\\\midrule
         \cmark&-&-&-&54.73&89.35 \\
         -&\cmark&-&-&54.81&89.29 \\
         -&-&\cmark&-&55.07&89.43\\
         -&-&-&\cmark&54.54&89.27 \\\midrule
         \cmark&-&-&-&54.73&89.35 \\
         \cmark&\cmark&-&-&55.49&89.41 \\
         \cmark&\cmark&\cmark&-&\textbf{55.60}&\textbf{89.44} \\
         \cmark&\cmark&\cmark&\cmark&55.38&89.35 \\\bottomrule
    \end{tabular}
    
    \label{tab:windowsizeselection}
\end{table}

\begin{table}[ht]
    \centering
    \footnotesize
    \setlength\tabcolsep{2pt}
    \caption{Ablation study on the impact of Adaptive Scale Weighting Module (ASWM) on the SMG dataset.}
    \begin{tabular}{ccc}
    \toprule
         \textbf{Multiscale aggregation type}& \textbf{Acc. Top-1(\%)}$\uparrow$& \textbf{Acc. Top-5(\%)}$\uparrow$ \\\midrule
         Average sum&55.60&89.44\\
         ASWM&\textbf{56.22}&\textbf{89.50} \\
         \bottomrule
    \end{tabular}
    \label{tab:fusion}
\end{table}

\begin{figure}[ht]
    \centering
    \footnotesize
    \setlength\tabcolsep{0pt}
    \begin{tabular}{cccc}
         \includegraphics[width=0.24\linewidth]{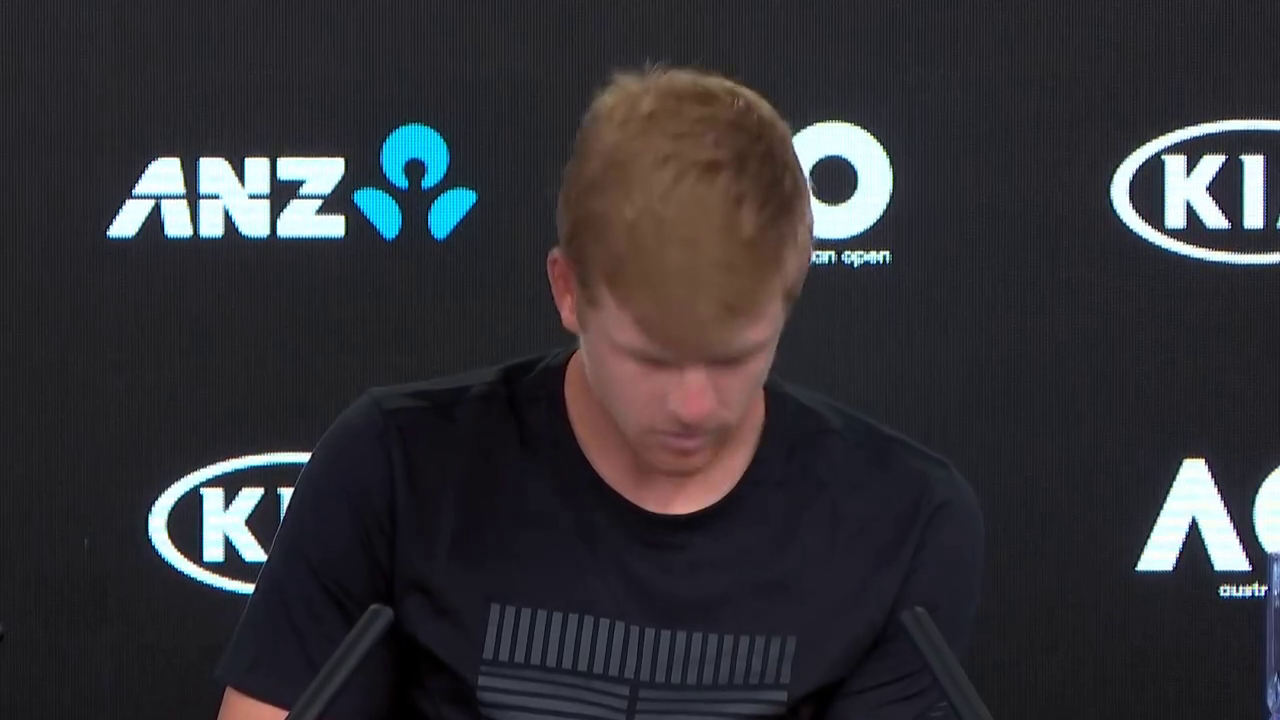}&\includegraphics[width=0.24\linewidth]{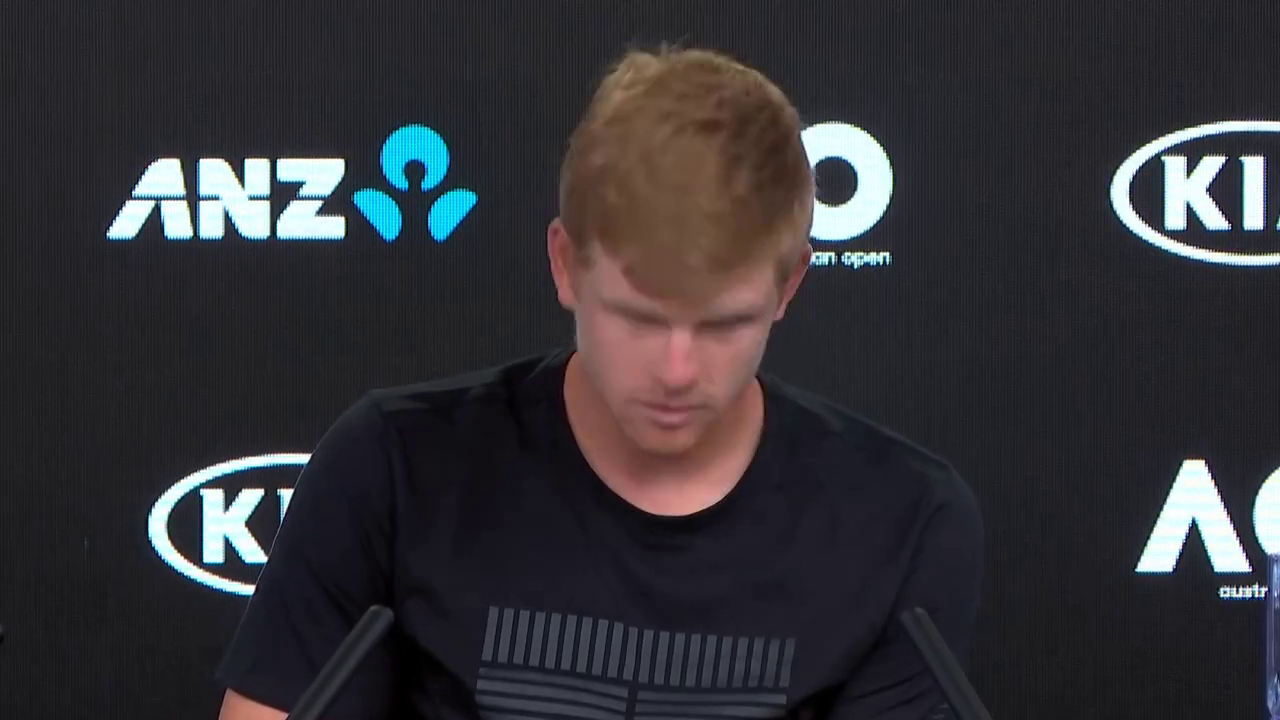}&\includegraphics[width=0.24\linewidth]{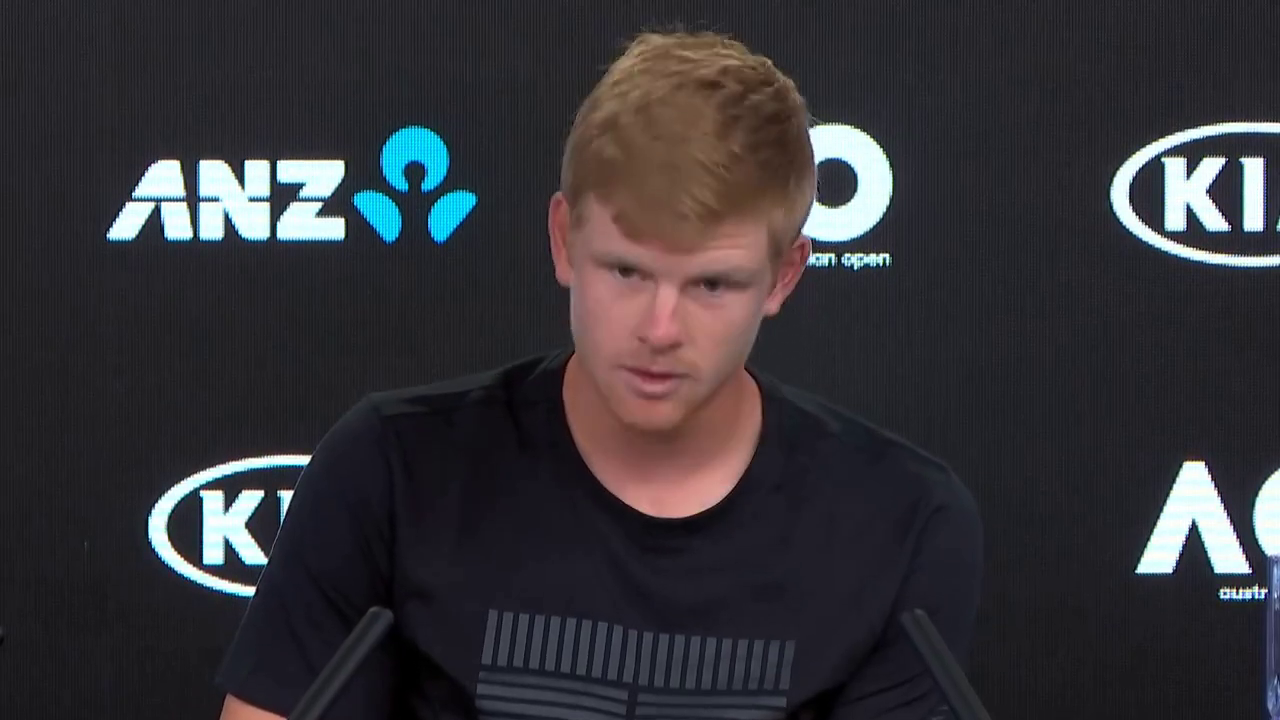}&\includegraphics[width=0.24\linewidth]{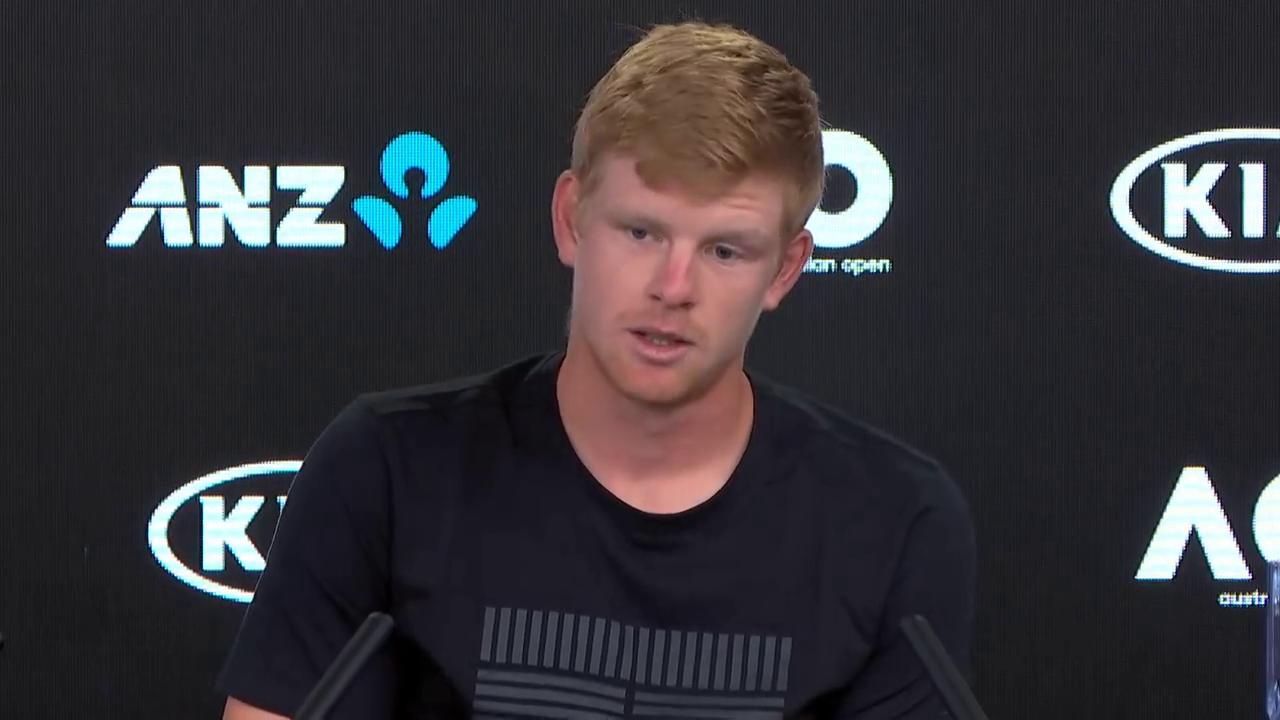}\\[-2pt]
         \multicolumn{4}{c}{\includegraphics[width=0.96\linewidth]{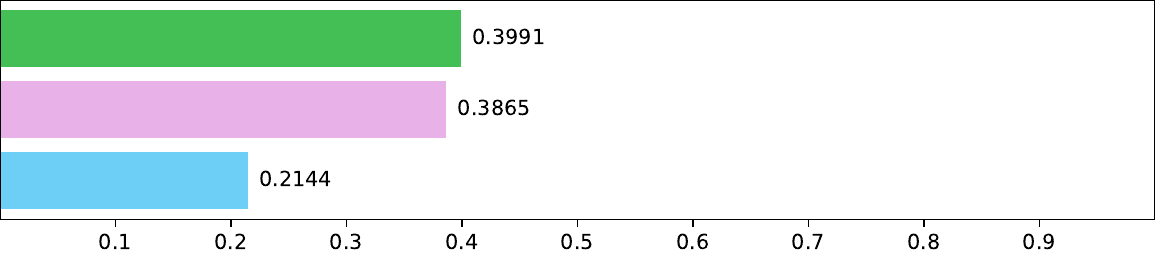}}\\
         \multicolumn{4}{c}{(a) Head up}\\
         \includegraphics[width=0.24\linewidth]{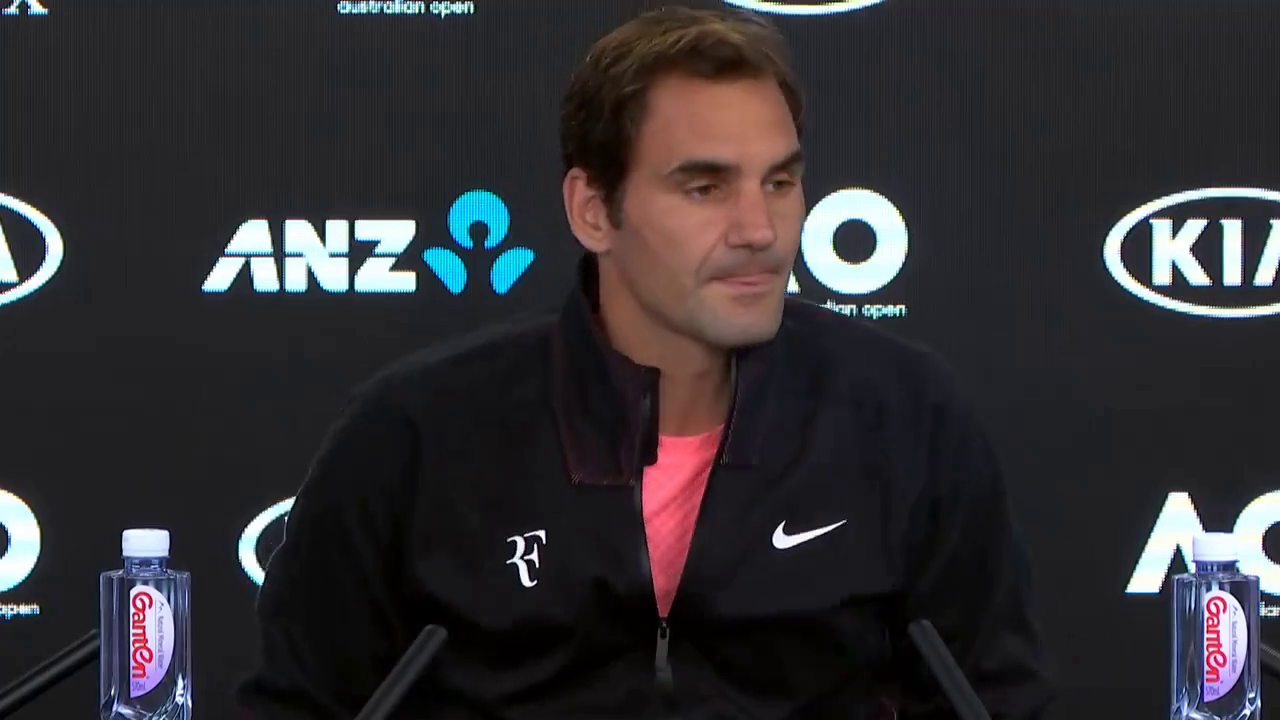}&\includegraphics[width=0.24\linewidth]{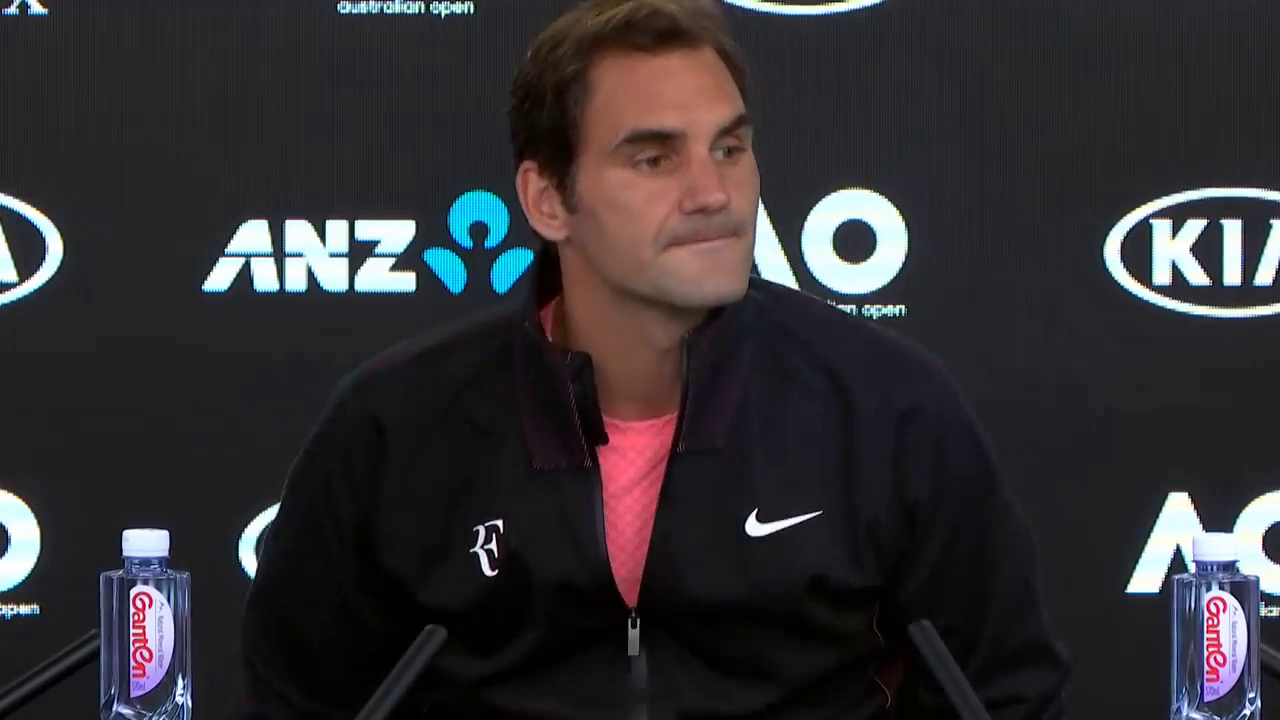}&\includegraphics[width=0.24\linewidth]{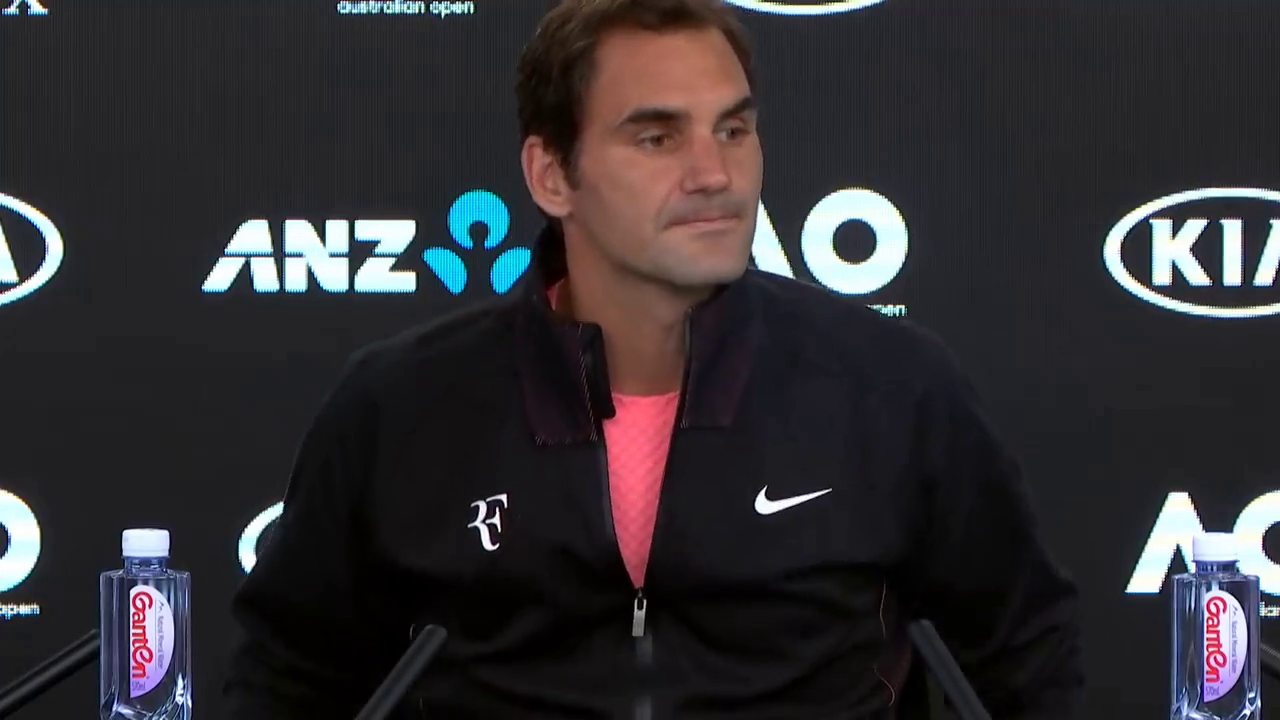}&\includegraphics[width=0.24\linewidth]{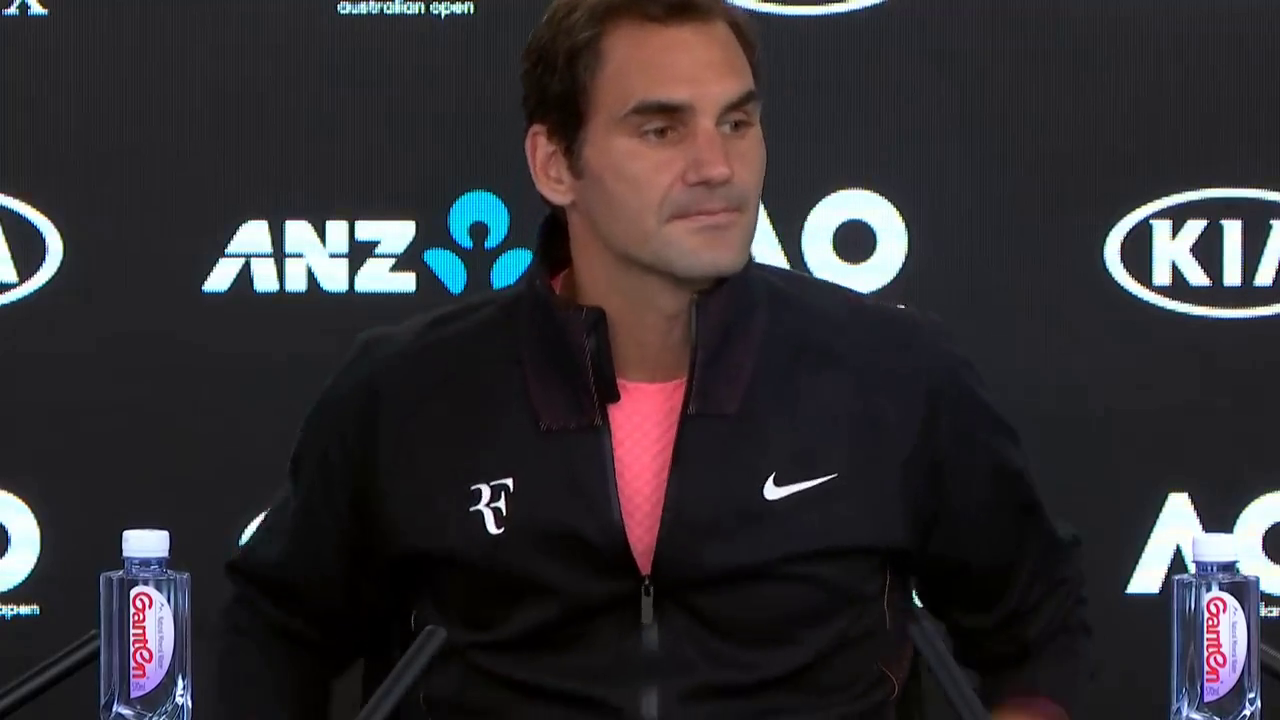}\\[-2pt]
         \multicolumn{4}{c}{\includegraphics[width=0.96\linewidth]{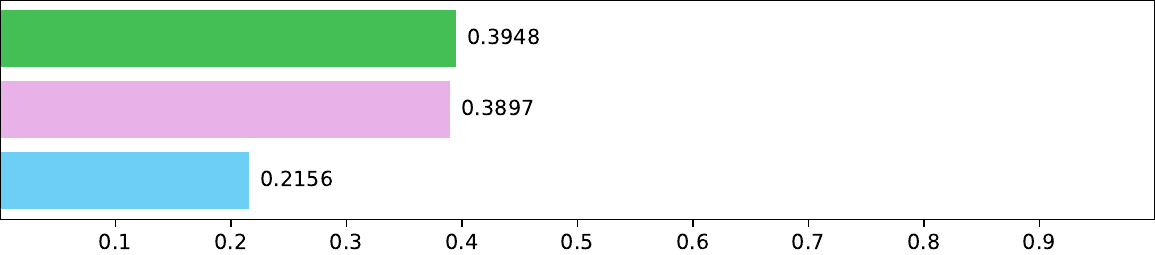}}\\
         \multicolumn{4}{c}{(b) Sit straightly}\\
         \includegraphics[width=0.24\linewidth]{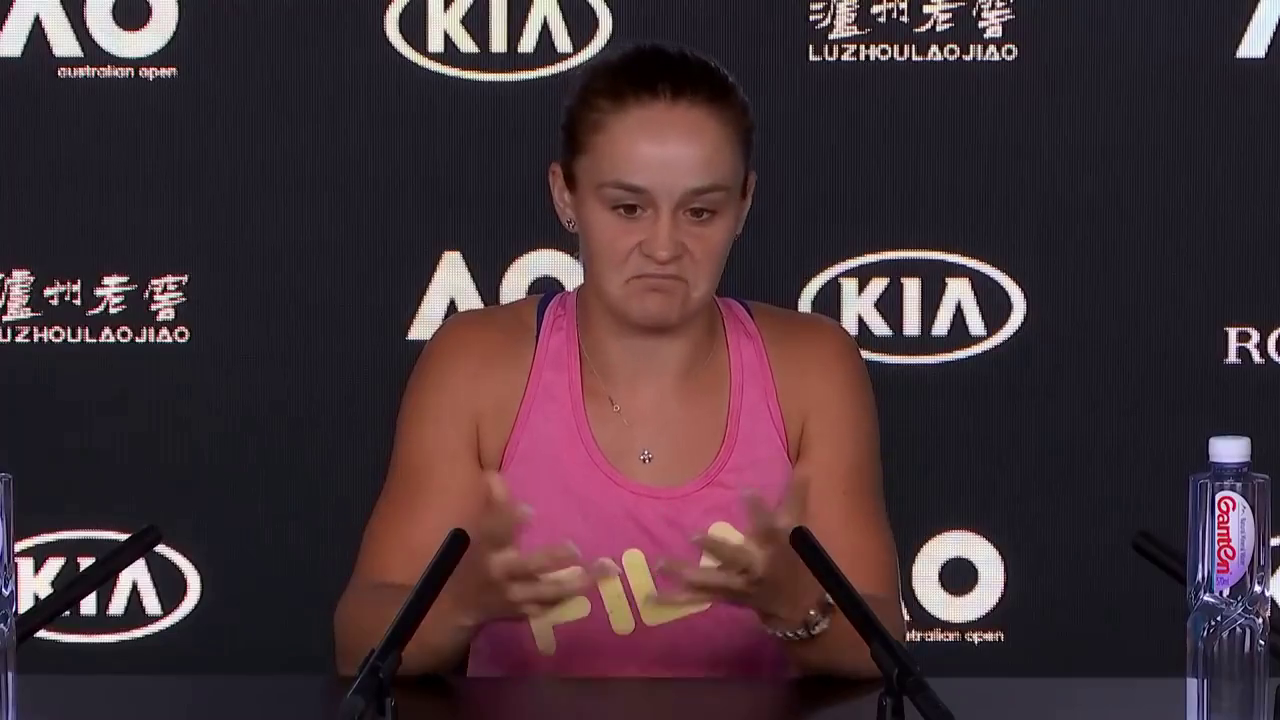}&\includegraphics[width=0.24\linewidth]{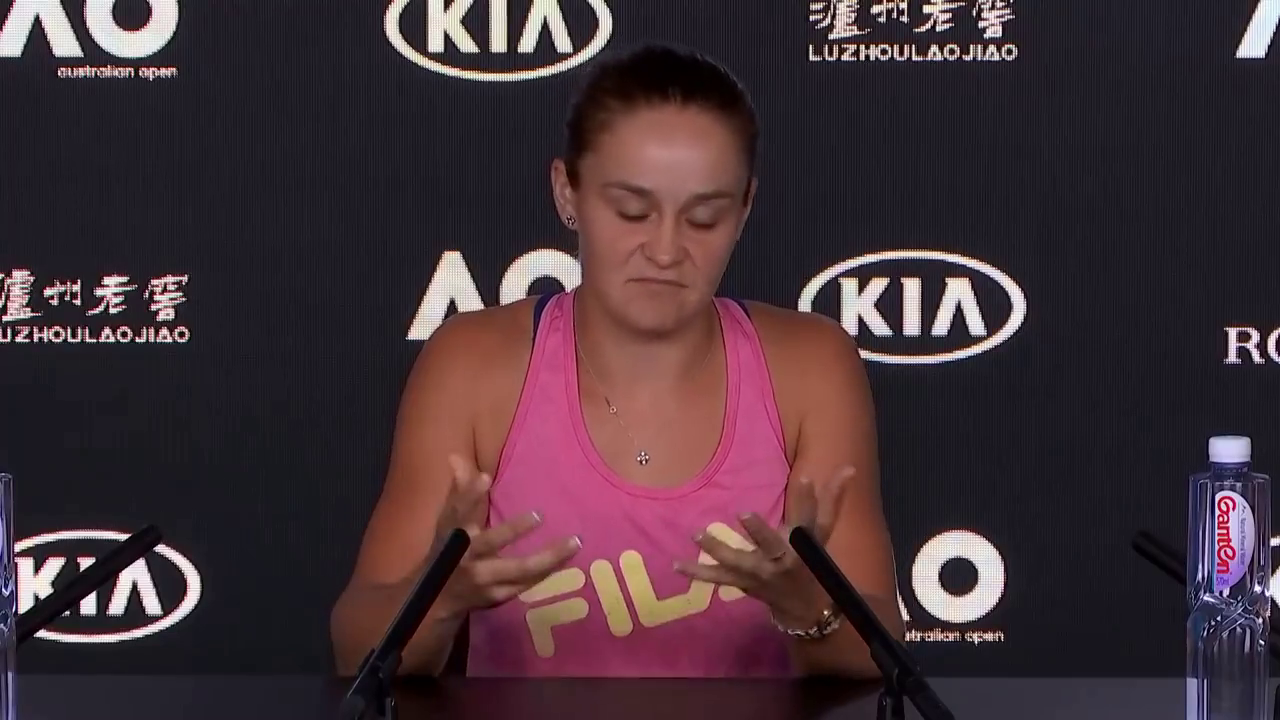}&\includegraphics[width=0.24\linewidth]{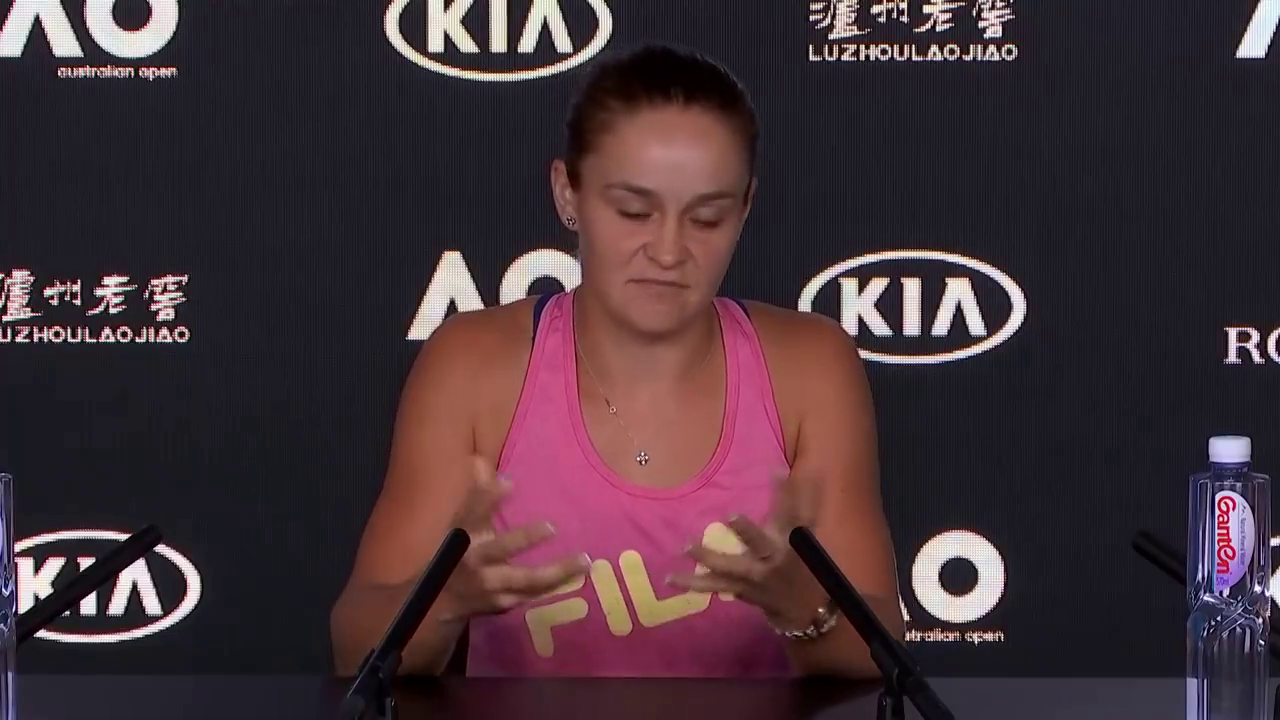}&\includegraphics[width=0.24\linewidth]{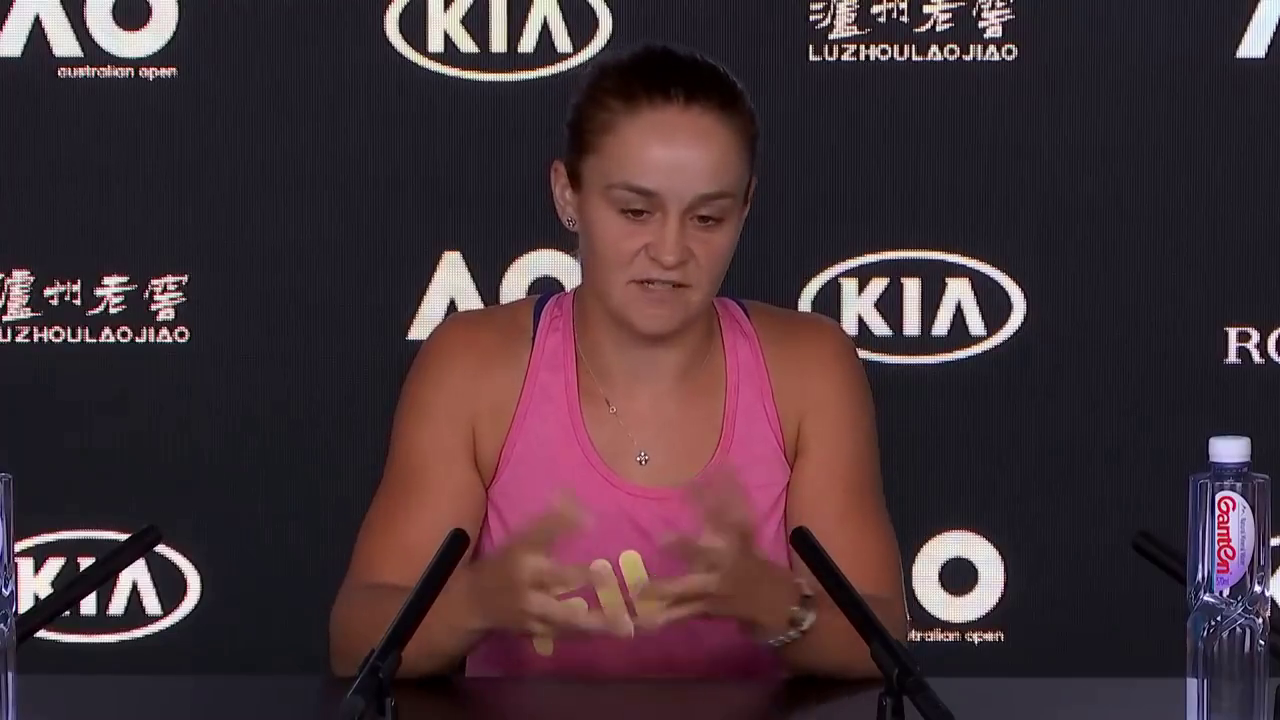}\\[-2pt]
         \multicolumn{4}{c}{\includegraphics[width=0.96\linewidth]{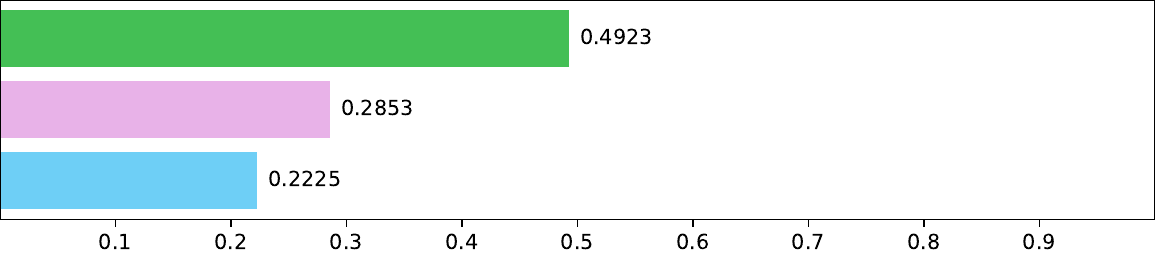}}\\
         \multicolumn{4}{c}{(c) Illustrative hand gestures}\\
         \includegraphics[width=0.24\linewidth]{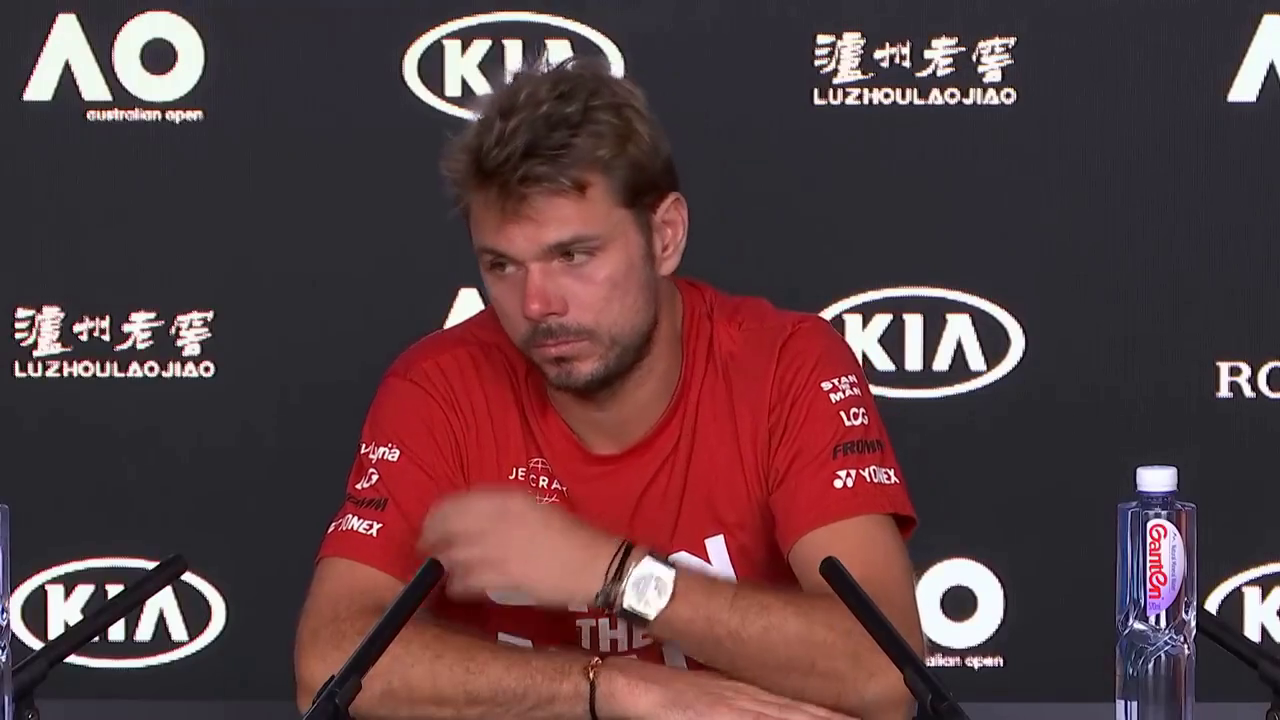}&\includegraphics[width=0.24\linewidth]{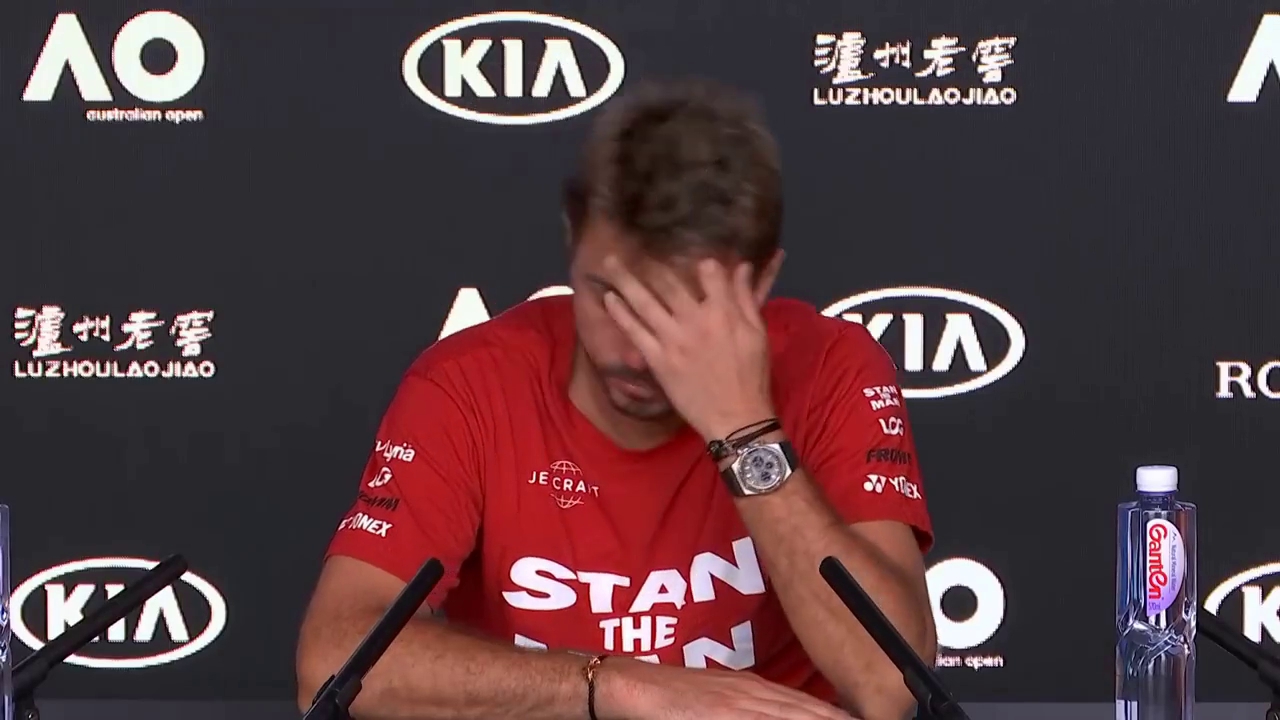}&\includegraphics[width=0.24\linewidth]{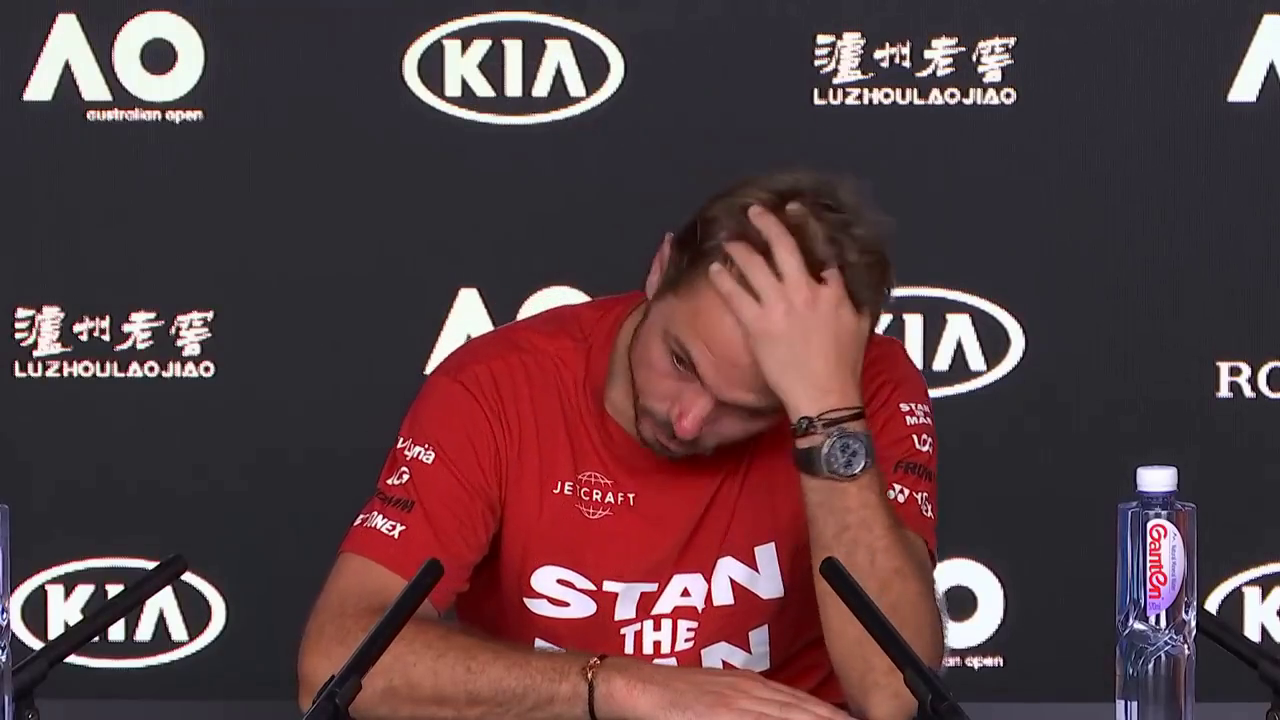}&\includegraphics[width=0.24\linewidth]{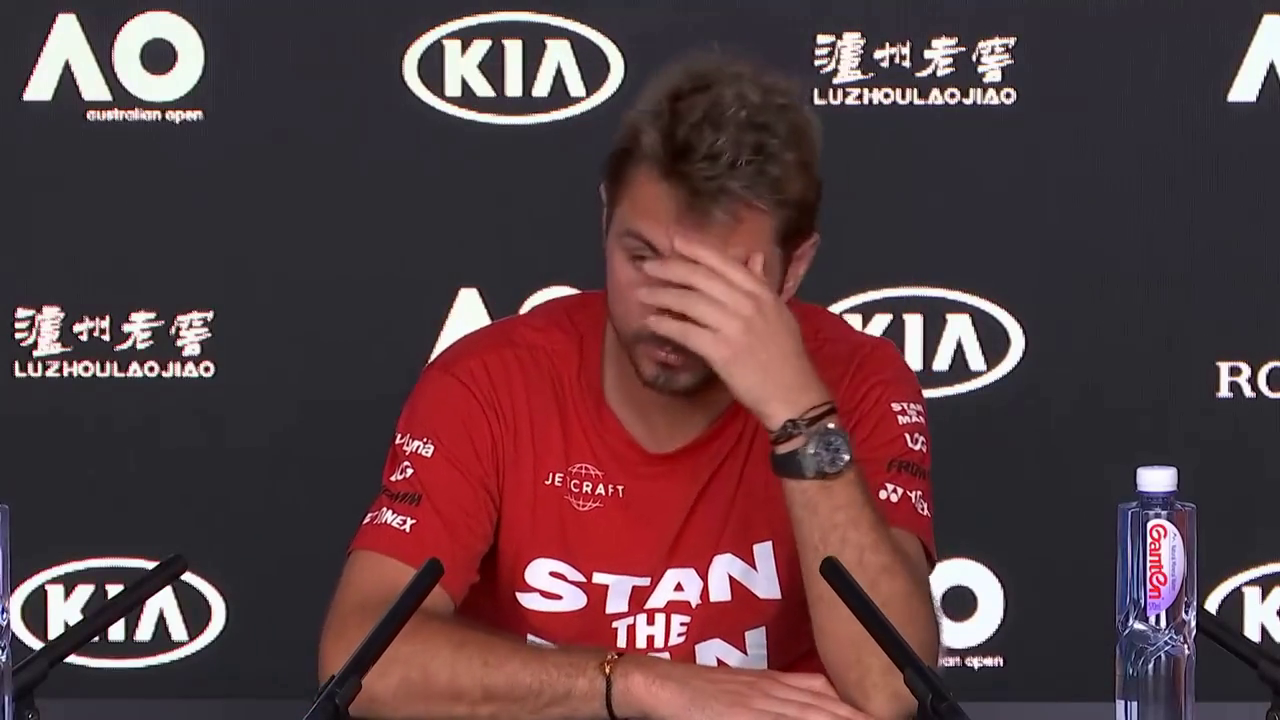}\\[-2pt]
         \multicolumn{4}{c}{\includegraphics[width=0.96\linewidth]{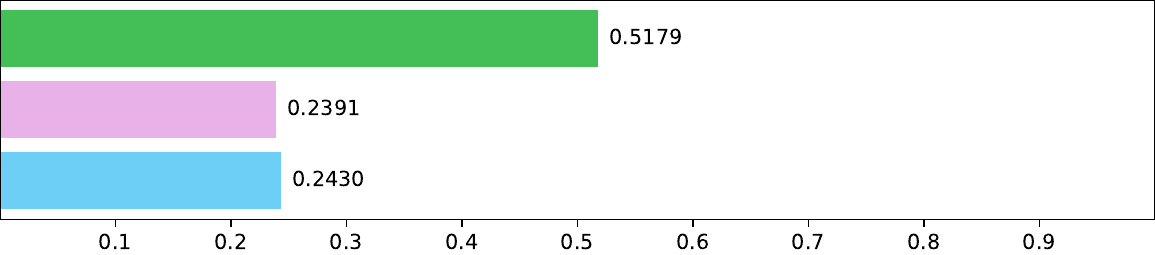}}\\
         \multicolumn{4}{c}{(d) Touching or scratching head}\\
    \end{tabular}
    \caption{Visualization of learned attention weights of ASWM over different fusion scales (\textcolor{skyblue}{$win@3\times3\times3$}, \textcolor{lavenderpink}{$win@5\times5\times5$}, \textcolor{leafgreen}{$win@7\times7\times7$}) for different micro-gestures.}
    \label{fig:ASWM}
\end{figure}

\textbf{Does adaptive scale weighting help?} We conduct an ablation study to evaluate the effect of the proposed ASWM, which dynamically assigns different weights to each motion-aware state fusion scale. As shown in Table~\ref{tab:fusion}, replacing ASWM with a simple average sum strategy (i.e., equal weights across $win@3\times3\times3$, $win@5\times5\times5$, and $win@7\times7\times7$ branches) leads to a drop in accuracy by 0.61\% on Top-1. This indicates that learning to adaptively weight different temporal-spatial scales is beneficial. \textcolor{black}{We also qualitatively analyze average attention weights across three window sizes and various MG types.} As shown in Figure~\ref{fig:ASWM}, we observe that larger window sizes (e.g., $Win@7\times7\times7$) tend to receive higher attention scores. The attention for \textit{"Touching or scratching head"} increases from 0.24 ($Win@3\times3\times3$) to 0.52 ($Win@7\times7\times7$), \textcolor{black}{indicating that smaller receptive fields fail to capture the full temporal-spatial context of this MG.} Similarly, \textit{"Illustrative hand gestures"} shows a consistent rise in attention with larger windows, which aligns with its inherently dynamic nature involving broad and continuous hand movements. In contrast, some MGs, such as \textit{"Sit straightly"} or \textit{"Head up"} exhibit a slightly increased attention across window sizes (from ~0.21 to ~0.39–0.40), suggesting that while longer context still improves modeling, these MGs \textcolor{black}{do not require as wide a spatiotemporal perception.} This further confirms the necessity of the proposed multiscale modeling strategies in the MGR task. 

In summary, our ablation experiments demonstrate that: 1) Different scan \textcolor{black}{strategies} of Mamba alone offers little gain, as all scans yield around 53.54\% Top-1 accuracy; 2) Motion-aware state fusion (MCFM + CFD) is the primary driver of improvement, boosting Top-1 accuracy from 53.18\% to 54.04\% even with a single scale ($win@3\times3\times3$); 3) Multiscale representations further enhance performance, with the optimal configuration using $Win@3\times3\times3$, $Win@5\times5\times5$, and $Win@7\times7\times7$ branches (55.60\% Top-1), while adding an excessively large 9×9×9 slightly degrades accuracy; 4) Adaptive fusion via ASWM outperforms uniform averaging, \textcolor{black}{confirming the benefit of dynamically weighting scales according to various MGs.}

\begin{table*}[t]
    \centering
    \footnotesize
    \setlength\tabcolsep{4pt}
    \caption{\textcolor{black}{Comparison the proposed \textit{MSF-Mamba} with state-of-the-art methods on iMiGUE and SMG dataset. \textcolor{lightgray}{Lightgray} represents the results reported in the iMiGUE~\cite{liu2021imigue} and the SMG~\cite{chen2023smg} paper, while \textcolor{black}{black} represents our reproduced results under the same training setting (e.g., number of frames, data augmentation).}}
    \begin{tabular}{lcccccc}
    \toprule
         \multirow{2}{*}{\textbf{Method}}&\multirow{2}{*}{\textbf{Model type}}&\multirow{2}{*}{\textbf{Model size (M)}}&\multicolumn{2}{c}{\textbf{iMiGUE}}&\multicolumn{2}{c}{\textbf{SMG}}\\
         &&&\textbf{Acc. Top-1(\%)$\uparrow$}&\textbf{Acc. Top-5(\%)$\uparrow$}&\textbf{Acc. Top-1(\%)$\uparrow$}&\textbf{Acc. Top-5(\%)$\uparrow$}\\
         \midrule
        \rowcolor{lightgray!20}
        C3D~\cite{tran2015learning} & CNN&-  & \textcolor{lightgray}{20.32} & \textcolor{lightgray}{55.31} & \textcolor{lightgray}{45.90} & \textcolor{lightgray}{79.18} \\
        \rowcolor{lightgray!20}
        I3D~\cite{carreira2017quo} & CNN&-  & \textcolor{lightgray}{34.96} & \textcolor{lightgray}{63.69} & \textcolor{lightgray}{35.08} & \textcolor{lightgray}{85.90} \\
        \rowcolor{lightgray!20}
        TSN~\cite{wang2016temporal} & CNN&-  & \textcolor{lightgray}{51.54} & \textcolor{lightgray}{85.42} & \textcolor{lightgray}{50.49} & \textcolor{lightgray}{82.13} \\
        \rowcolor{lightgray!20}
        TSM~\cite{lin2019tsm} & CNN&- & \textcolor{lightgray}{61.10} & \textcolor{lightgray}{91.24} & \textcolor{lightgray}{58.69} & \textcolor{lightgray}{83.93} \\
        \rowcolor{lightgray!20}
        TRN~\cite{zhou2018temporal} & CNN&- & \textcolor{lightgray}{55.24} & \textcolor{lightgray}{89.17} & \textcolor{lightgray}{59.51} & \textcolor{lightgray}{88.53} \\
         \midrule
         \multicolumn{7}{c}{\textit{CNN-based}}\\
         C3D~\cite{tran2015learning} & CNN&78.13 &30.13&58.84& 36.07 & 76.85\\
         I3D~\cite{carreira2017quo} & CNN &27.29&35.08&62.70&28.36 & 68.20\\TSN~\cite{wang2016temporal} & CNN&23.57 &51.31&84.42& 39.59 & 78.03 \\TSM~\cite{lin2019tsm} & CNN &23.57& 56.23&88.39&47.10 & 79.23 \\
         MA-Net~\cite{guo2024benchmarking}& CNN&43.23 &58.08&84.09& 48.69 & 79.51 \\
         \midrule
         \multicolumn{7}{c}{\textit{Transformer-based}}\\Timesformer~\cite{bertasius2021space} & Transformer&86.11 &50.66&84.76& 47.89 & 86.39 \\
         VSwin~\cite{liu2022video} & Transformer&49.53 &60.11 &86.62& 48.03 & 89.18  \\
         UniformerV2~\cite{li2023uniformerv2} & Transformer &49.80 &59.91&90.30& 46.07 & 87.38 \\\hline
         \multicolumn{7}{c}{\textit{Mamba-based}}\\
         VideoMamba~\cite{li2025videomamba}&SSM&26.00&58.13&90.95&53.28&87.54\\
         \rowcolor{blue!20}MSF-Mamba-T&SSM&8.90&57.30&89.60&50.57&87.31\\
         \rowcolor{blue!20}MSF-Mamba-T$^{+}$&SSM&24.93&58.33&90.53&52.29&88.19\\
         \rowcolor{blue!20}MSF-Mamba-S&SSM&33.38&60.32&90.81&54.73&89.35\\
         \rowcolor{blue!20}MSF-Mamba-S$^{+}$&SSM&97.28&61.17&91.43&56.22&89.50\\
         \rowcolor{blue!20}MSF-Mamba-M&SSM&91.57&61.19&91.45&59.67&89.66\\
         \rowcolor{blue!20}MSF-Mamba-M$^{+}$&SSM&235.32&\textbf{62.98}&\textbf{92.28}&\textbf{60.13}&\textbf{89.83}\\
    \bottomrule
    \end{tabular}
    \label{tab:Compara}
\end{table*}

\subsection{Comparative Experiment}

\begin{table}[ht]
    \centering
    \footnotesize
    \setlength\tabcolsep{2pt}
    \caption{Architecture configurations of different MSF-Mamba$^{+}$ and MSF-Mamba model variants. All models use a patch size of 16×16.}
    \begin{tabular}{lccccc}\toprule
         \multirow{2}{*}{\textbf{Method}}&\multirow{2}{*}{\textbf{MCFM}}&\multirow{2}{*}{\textbf{ASWM}}&\textbf{Embedding}&\multirow{2}{*}{\textbf{\#Layers}}&\textbf{\#Params}  \\
         &&&\textbf{dimension}&&\textbf{(M)}\\\midrule
         MSF-Mamba-T&win@3&-&192&24&8.90\\
         MSF-Mamba-S&win@3&-&384&24&33.38\\
         MSF-Mamba-M&win@3&-&576&32&91.57\\
         MSF-Mamba-T$^{+}$&win@3,5,7&\cmark&192&24&24.93\\
         MSF-Mamba-S$^{+}$&win@3,5,7&\cmark&384&24&97.28\\
         MSF-Mamba-M$^{+}$&win@3,5,7&\cmark&576&32&235.32\\
         \bottomrule
    \end{tabular}
    \label{tab:model_config}
\end{table}

\textcolor{black}{To ensure a fair comparison} with existing models, we develop two versions of our method: 1) a full-featured model, referred to as MSF-Mamba$^{+}$; 2) a lightweight variant called MSF-Mamba, which uses only a single-scale $win@3\times3\times3$ state fusion branch and excludes ASWM. In addition, we follow the configuration of VideoMamba~\cite{li2025videomamba} to construct three model variants—Tiny, Small, and Middle by adjusting the embedding dimension and depth. All models use a patch size of $16\times16$. The detailed configurations are summarized in Table~\ref{tab:model_config}. 

To evaluate the effectiveness of our proposed method, we compare both the full-featured version MSF-Mamba$^+$ and the lightweight version MSF-Mamba with existing state-of-the-art (SOTA) MGR models across three categories: CNN-based, Transformer-based, and Mamba-based models. All baseline models are re-implemented using the MMAction2~\footnote{https://github.com/open-mmlab/mmaction2} framework under a standardized experimental setup, where 16 frames are uniformly sampled and consistent data augmentation strategies are applied. Table~\ref{tab:Compara} summarizes the results, where gray entries denote values reported in prior works~\cite{liu2021imigue, chen2023smg}, and black entries represent our reproduced results. CNN-based methods, such as C3D~\cite{tran2015learning}, I3D~\cite{carreira2017quo}, and MA-Net~\cite{guo2024benchmarking}, demonstrate moderate performance due to their limited ability to capture long-range dependencies. For instance, on the SMG dataset, C3D achieves only 36.07\% Top-1 accuracy, while MA-Net improves this to 48.69\%, yet still underperforms on fine-grained MG. Transformer-based models, including TimeSformer~\cite{bertasius2021space}, Video Swin~\cite{liu2022video}, and UniformerV2~\cite{li2023uniformerv2}, achieve higher accuracy due to their global attention mechanisms. However, their large parameter sizes and quadratic complexity make them computationally expensive. For example, UniformerV2 reaches 59.91\% on iMiGUE and 46.07\% on SMG with nearly 50M parameters. Mamba-based methods provide a promising trade-off between efficiency and performance. VideoMamba~\cite{li2025videomamba} achieves 58.13\% Top-1 accuracy on iMiGUE and 53.28\% on SMG with just 26M parameters. Our lightweight model MSF-Mamba introduces motion-aware state fusion to Mamba. MSF-Mamba-S achieves significantly better results, namely, 61.32\% Top-1 accuracy on iMiGUE and 54.73\% on SMG. It demonstrates the benefits of incorporating local spatiotemporal modeling into Mamba. Furthermore, the full model MSF-Mamba-S$^+$, which adds additional multiscale state fusion and attention components, further improves performance on SMG to 56.22\% Top-1 accuracy and 89.50\% Top-5 accuracy. \textcolor{black}{On the iMiGUE dataset, MSF-Mamba-S$^+$ achieves 61.17\% Top-1 and 91.43\% Top-5 accuracy. Finally, MSF-Mamba-M$^+$ achieves the best performance, with 62.98\% Top-1 accuracy on iMiGUE.} This result validates the effectiveness of our proposed design across different model scales.

\subsection{Efficiency Comparison}

\begin{figure}[t]
    \centering
    \includegraphics[width=0.85\linewidth]{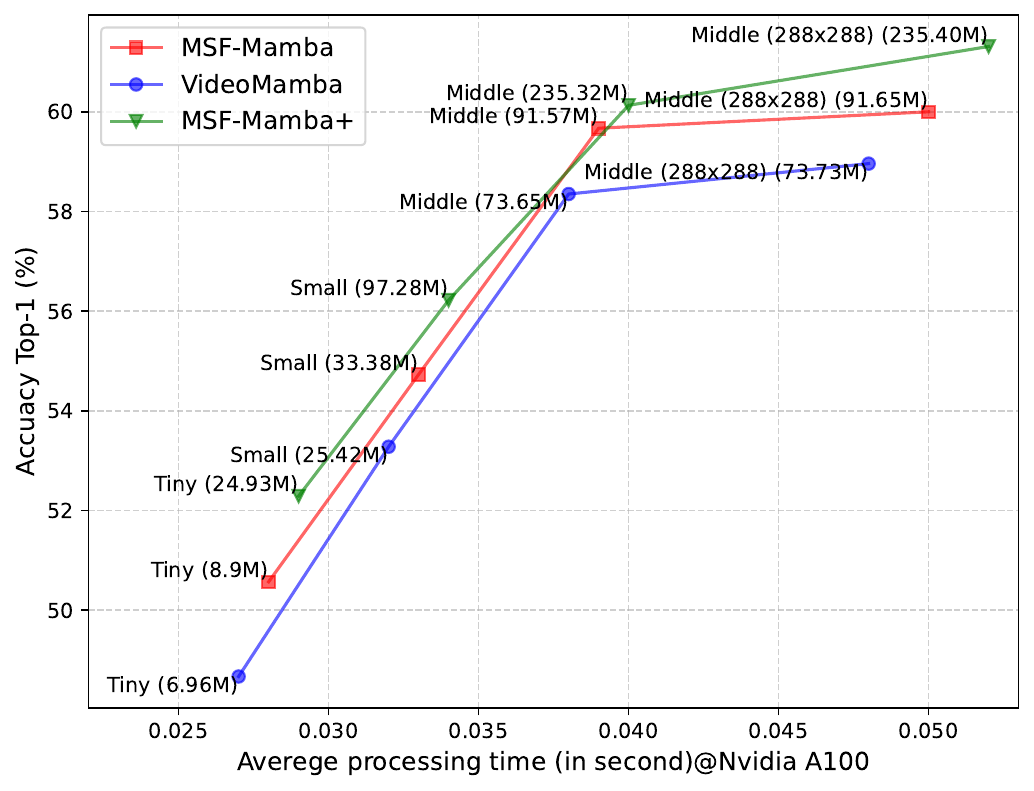}
    \caption{\textcolor{black}{Comparison between MSF-Mamba, MSF-Mamba$^{+}$, and VideoMamba at different model scales on the SMG dataset. The x-axis indicates average inference time per video (in seconds), and the y-axis shows Top-1 accuracy.}}
    \label{fig:compareonSMGmamba}
\end{figure}

\begin{figure}[t]
    \centering
    \includegraphics[width=0.85\linewidth]{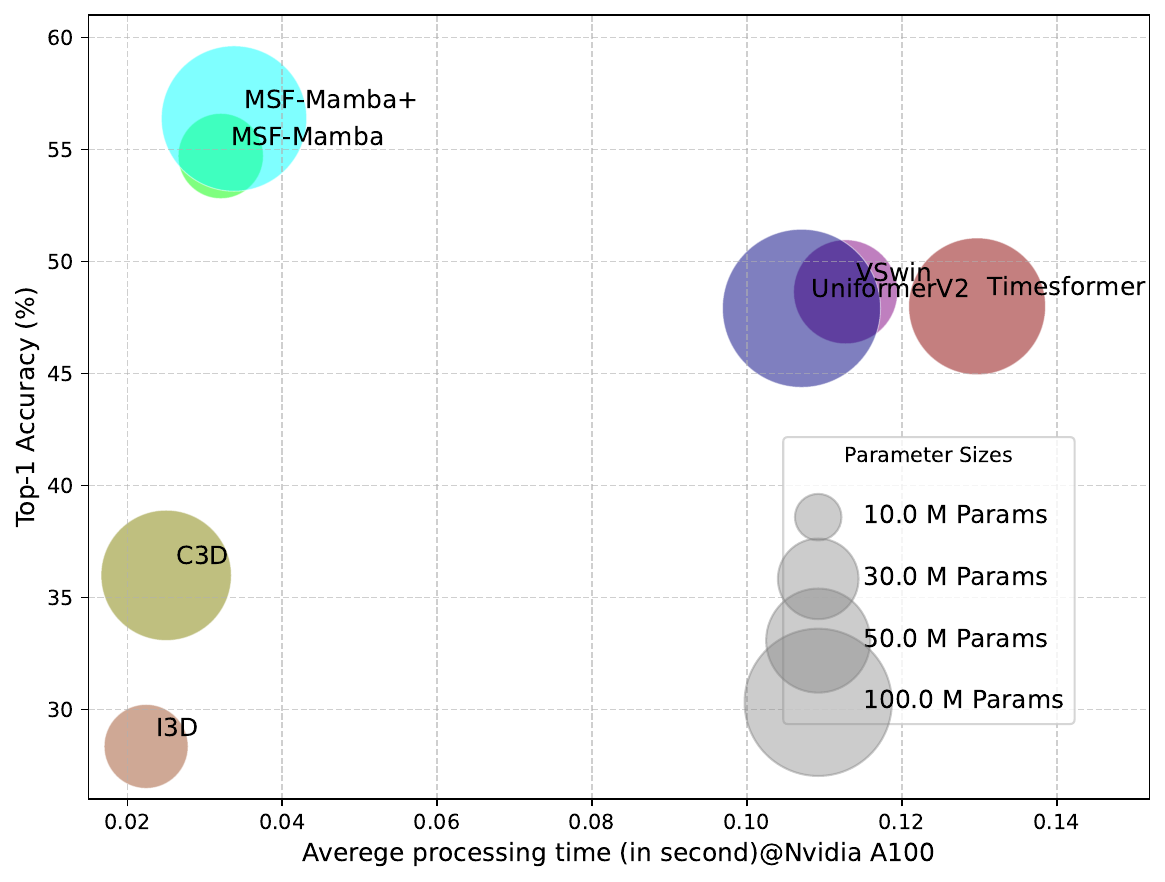}
    \caption{\textcolor{black}{Accuracy-efficiency-parameter trade-off comparison on the SMG dataset. Each bubble represents a method, where the x-axis indicates average inference time per video (in seconds), the y-axis shows Top-1 accuracy, and the bubble size denotes the number of parameters (in millions).}}
    \label{fig:compareonSMG}
\end{figure}

To further assess the efficiency of the proposed method, we conduct a comprehensive comparison in terms of recognition accuracy, inference latency, and model size. \textcolor{black}{First}, we compare MSF-Mamba, MSF-Mamba$^+$, and the baseline VideoMamba~\cite{li2025videomamba} across different model scales on the SMG dataset as shown in Figure~\ref{fig:compareonSMGmamba}. \textcolor{black}{The resolution of input MG videos is $224\times224$.} The $x$-axis denotes the average inference time per video (in seconds) on an NVIDIA A100 GPU, and the $y$-axis shows the corresponding Top-1 accuracy. As shown in Figure~\ref{fig:compareonSMGmamba}, MSF-Mamba consistently outperforms VideoMamba across all scales with only a marginal increase in processing time. The full model MSF-Mamba$^+$ delivers the highest performance at a slightly higher cost. These results demonstrate that our motion-aware and multiscale fusion modules introduce minimal latency while offering significant performance improvements. \textcolor{black}{Notably, although MSF-Mamba$^+$ incorporates more complex modules (e.g., multiscale state fusion and ASWM) and additional parameters, its core remains the Mamba/SSM architecture.} The key advantage of this type of model is that the inference complexity is linear $\mathcal{O}(n)$. Even though MSF-Mamba$^+$ has more parameters, \textcolor{black}{the inference latency does not increase linearly.} Figure~\ref{fig:compareonSMG} visualizes the trade-off among accuracy, latency, and model size for representative methods. Each bubble’s position represents inference time and Top-1 accuracy, while its size indicates the number of parameters. Compared to Transformer-based methods such as TimeSformer~\cite{bertasius2021space}, Video Swin~\cite{liu2022video}, and UniformerV2~\cite{li2023uniformerv2}, our MSF-Mamba and MSF-Mamba$^+$ offer significantly better accuracy-efficiency trade-offs. Transformer models typically require \textcolor{black}{large computational cost}, while our methods achieve superior accuracy at lower computational costs.


\section{Limitations}

\begin{figure}[h]
    \centering
    \footnotesize
    \setlength\tabcolsep{0pt}
    \begin{tabular}{ccc}
         \includegraphics[width=0.4\linewidth]{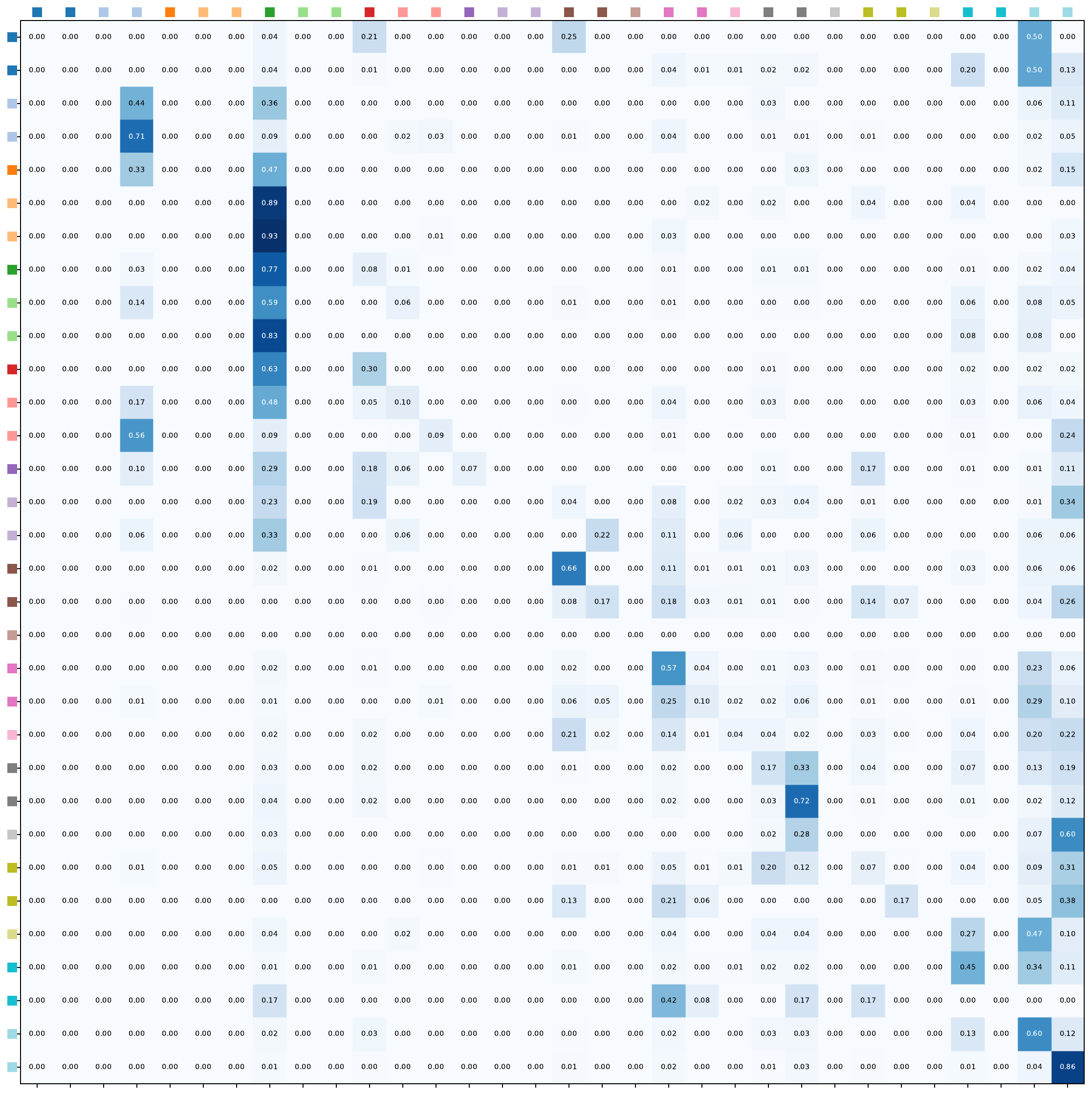}&\includegraphics[width=0.4\linewidth]{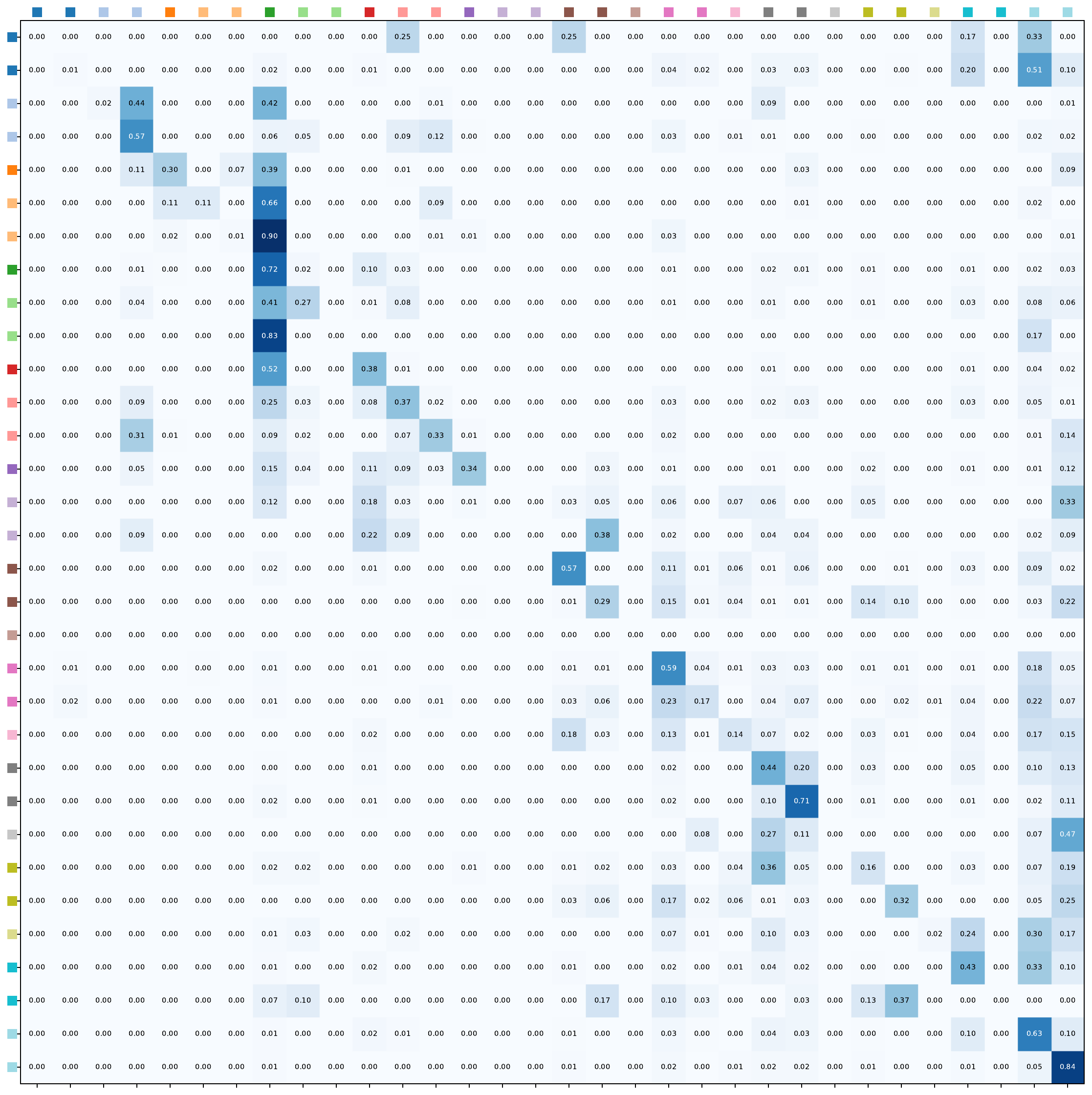}&\includegraphics[width=0.143\linewidth]{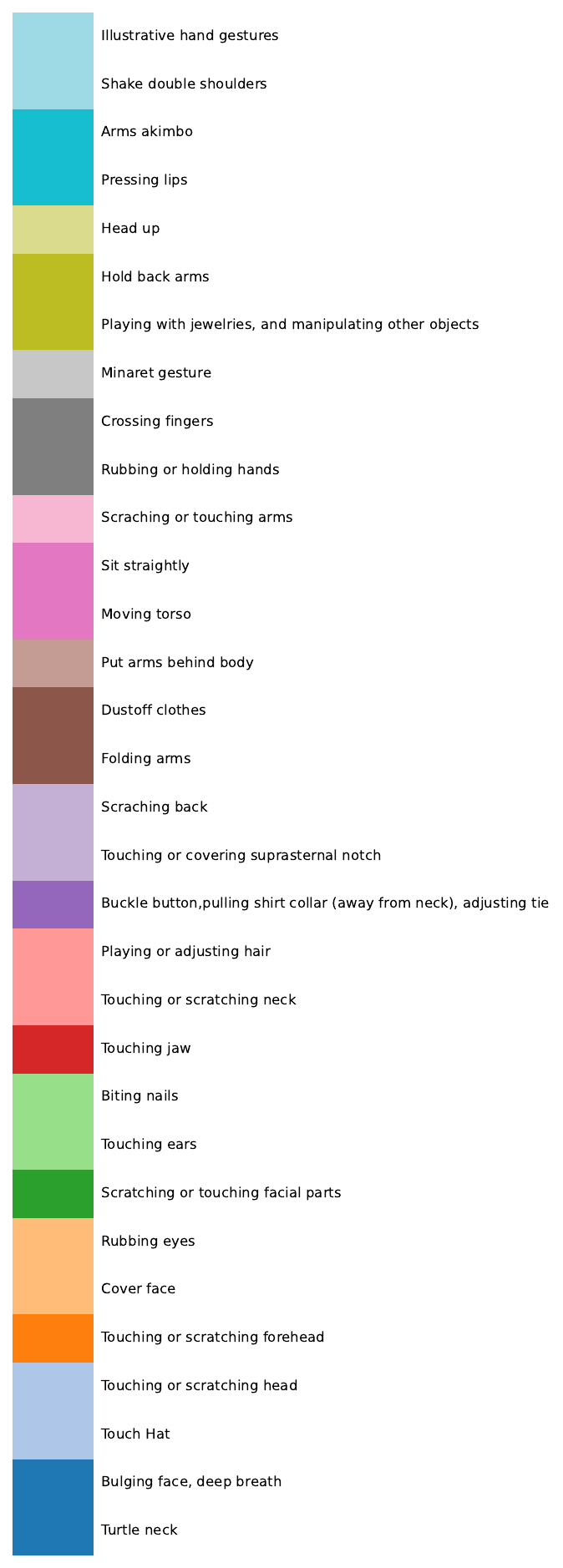}\\
         (a)&(b)&(c)\\
    \end{tabular}
    \caption{Comparison of confusion matrices on the iMiGUE dataset: (a) Confusion matrix of the baseline VideoMamba model; (b) Confusion matrix of our proposed MSF-Mamba$^{+}$; (c) Vertical legend for micro-gesture (MG) class names. View digitally and zoom in may be better.}
    \label{fig:confusionimigue}
\end{figure}

\begin{figure}[h]
    \centering
    \setlength\tabcolsep{0pt}
    \begin{tabular}{ccc}
        
         \includegraphics[width=0.4\linewidth]{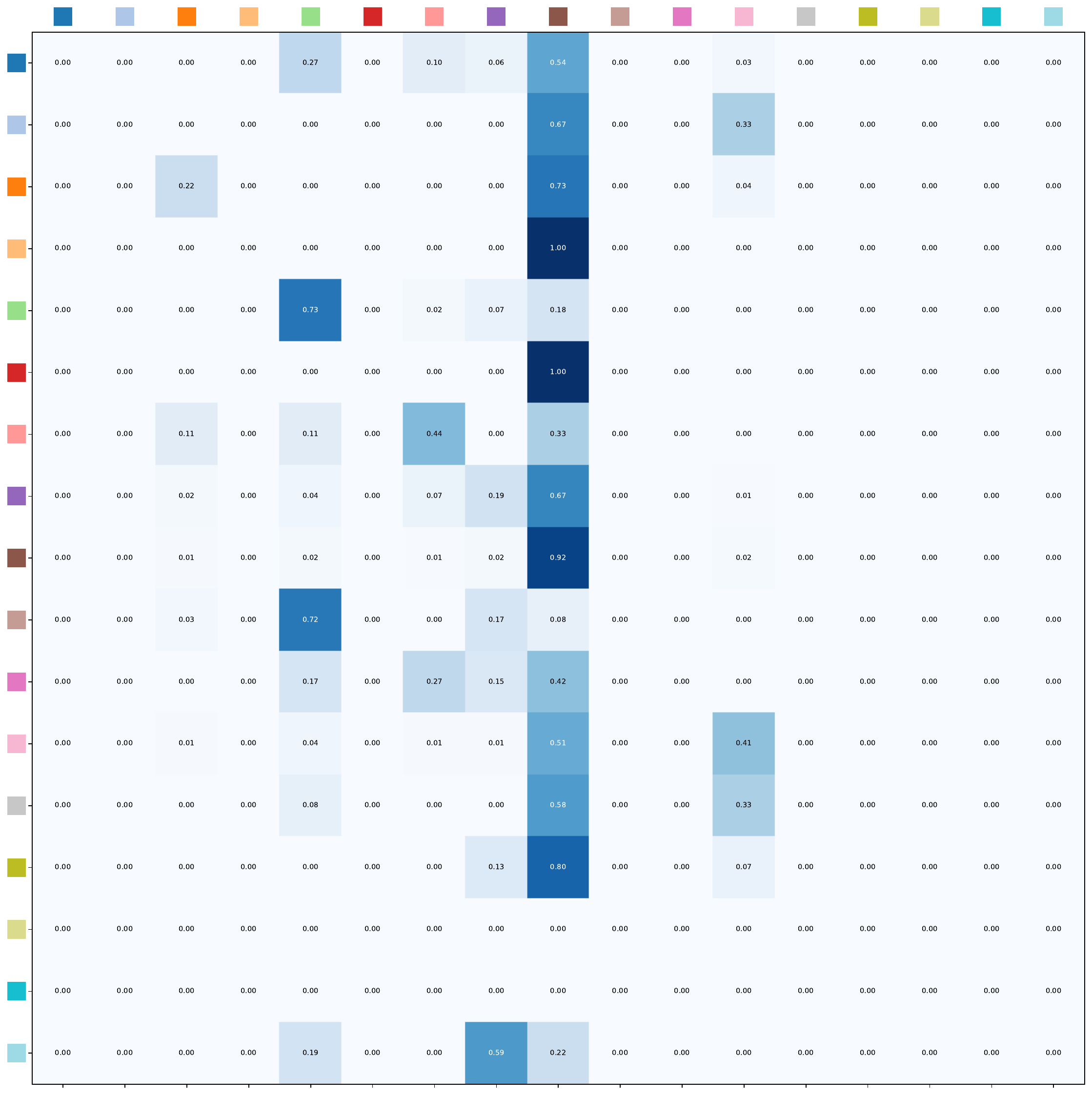}&\includegraphics[width=0.4\linewidth]{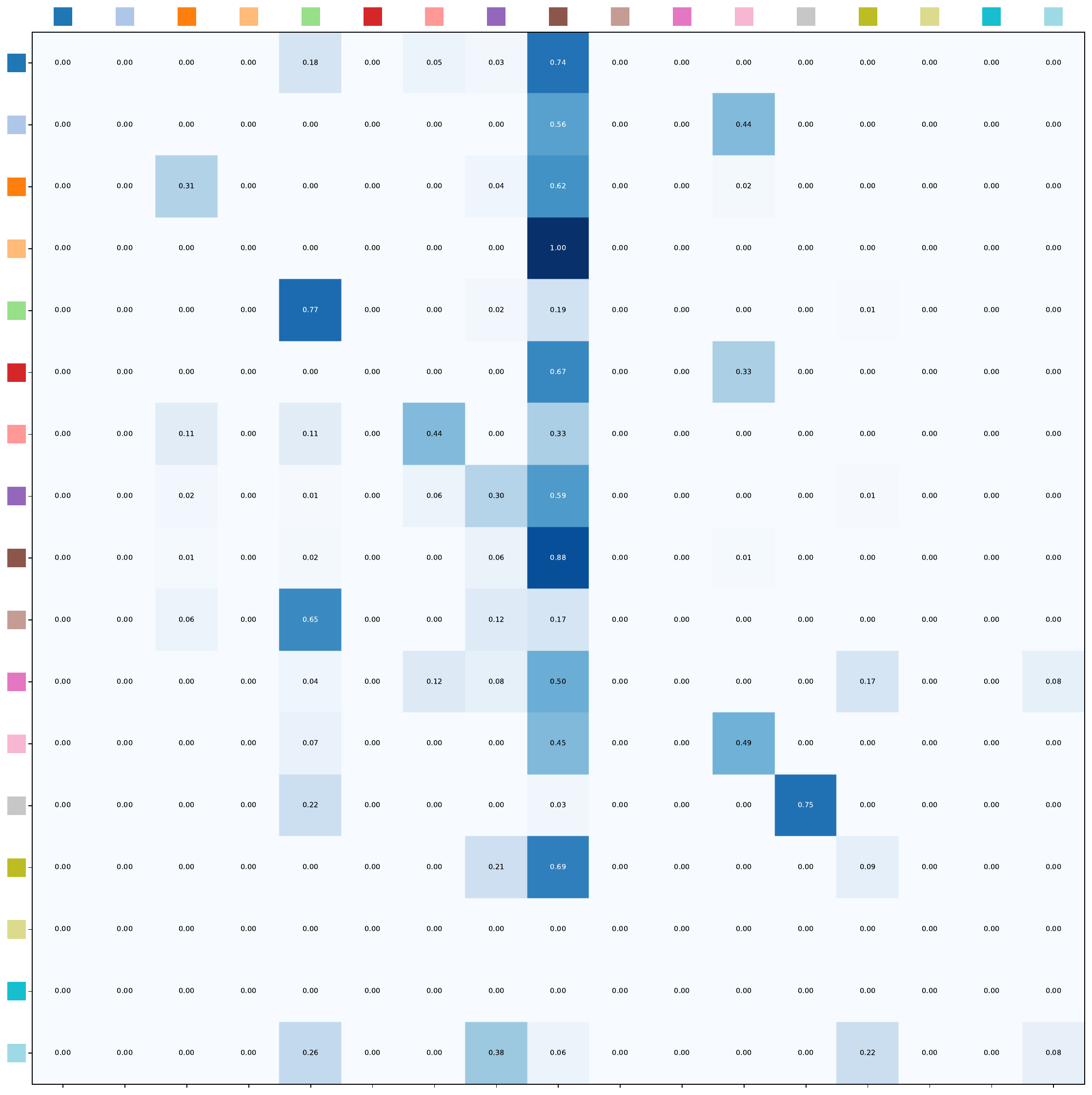}&\includegraphics[width=0.155\linewidth]{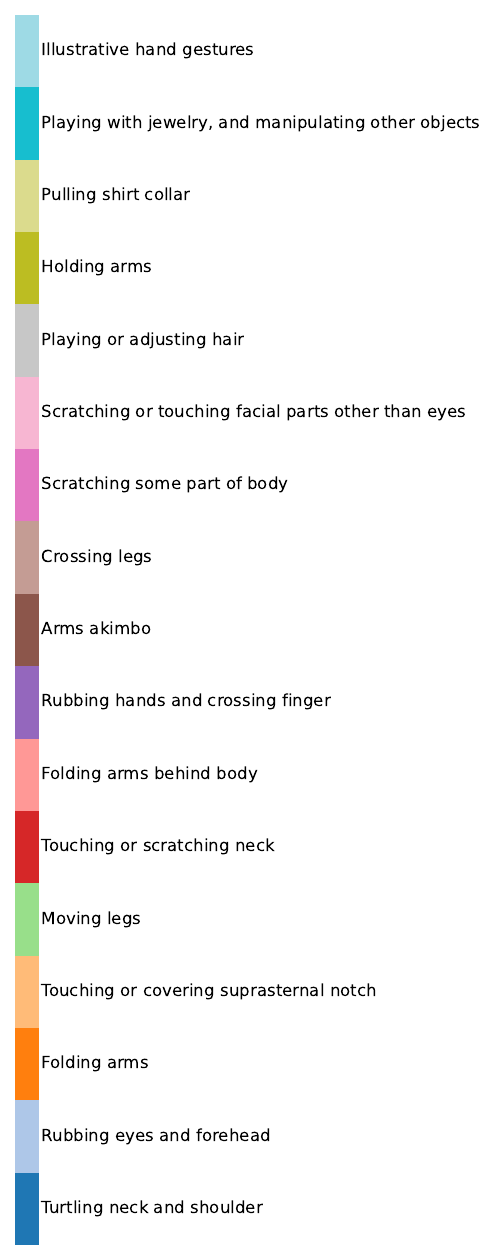}  \\
         (a)&(b)&(c)\\
    \end{tabular}
    \caption{Comparison of confusion matrices on the SMG dataset: (a) Confusion matrix of the baseline VideoMamba model; (b) Confusion matrix of our proposed MSF-Mamba$^{+}$; (c) Vertical legend for micro-gesture (MG) class names. View digitally and zoom in may be better.}
    \label{fig:confusionsmg}
\end{figure}

Despite achieving SoTA performance compared to existing CNN-, Transformer-, and Mamba-based methods, our proposed MSF-Mamba and MSF-Mamba$^{+}$ still face limitations. For example, the overall recognition accuracy on the iMiGUE dataset remains around 60\% Top-1, which suggests that MGR remains a highly challenging task. This relatively modest performance indicates that subtle differences between fine-grained MGs are still difficult to resolve, such as low motion amplitude or appearance similarities. To further investigate these limitations, we present a comparison of confusion matrices between VideoMamba and our MSF-Mamba-S$^{+}$ in Figure~\ref{fig:confusionimigue} and Figure~\ref{fig:confusionsmg}. MSF-Mamba-S$^{+}$ reduces several types of misclassifications. For instance, it improves the discrimination between \textit{"Scratching or touching facial parts"} and \textit{Touching ears} on the iMiGUE dataset and 
\textit{"Touching or scratching neck"} and \textit{"Arms akimbo"} on the SMG dataset. \textcolor{black}{However, visually similar MGs such as \textit{"Rubbing eyes"} vs. \textit{"Scratching or touching facial parts"} continue to challenge the model. Even with our motion-aware enhancements, the purely spatiotemporal information may not be sufficient to resolve these subtle differences. The future research could explore incorporating domain-specific priors, such as body part segmentation and hand pose estimation.} This could be implemented in the following ways: 1) Multi-task learning: body part segmentation or keypoint detection is learned jointly with MGR; 2) Body part guided attention mechanisms: incorporate explicit part location cues (e.g., pose keypoints) into the model via learned spatial attention. The body part-conditioned module can help the model focus on relevant regions.

\section{Conclusion}

In this paper, we introduce MSF-Mamba, a novel Mamba-based architecture for micro-gesture recognition (MGR). \textcolor{black}{Unlike the vanilla Mamba, which lacks spatiotemporal awareness due to its purely sequential state-update mechanism, MSF-Mamba explicitly models local motion patterns through a motion-aware central-frame-difference state-fusion module (MCFM).} To further improve the representation of multiscale gesture dynamics, we propose an enhanced version, MSF-Mamba$^+$, which incorporates multiscale state fusion and a dynamic adaptive scale weighting module (ASWM). Extensive experiments on two public MGR datasets, iMiGUE and SMG, demonstrate that MSF-Mamba and MSF-Mamba$^+$ significantly outperform existing CNN-, Transformer-, and SSM-based methods while maintaining high computational efficiency. \textcolor{black}{Notably, even the lightweight MSF-Mamba achieves substantial performance gains over VideoMamba with minimal overhead, while the full MSF-Mamba$^+$ further improves accuracy.} These results highlight the importance of integrating motion-aware local spatiotemporal modeling into efficient sequence models for recognizing subtle micro-gestures.

\bibliographystyle{IEEEtran}
\bibliography{mybibliography}

\ifCLASSOPTIONcaptionsoff
  \newpage
\fi



%

%
\begin{IEEEbiography}[{\includegraphics[width=1in,height=1.25in,clip,keepaspectratio]{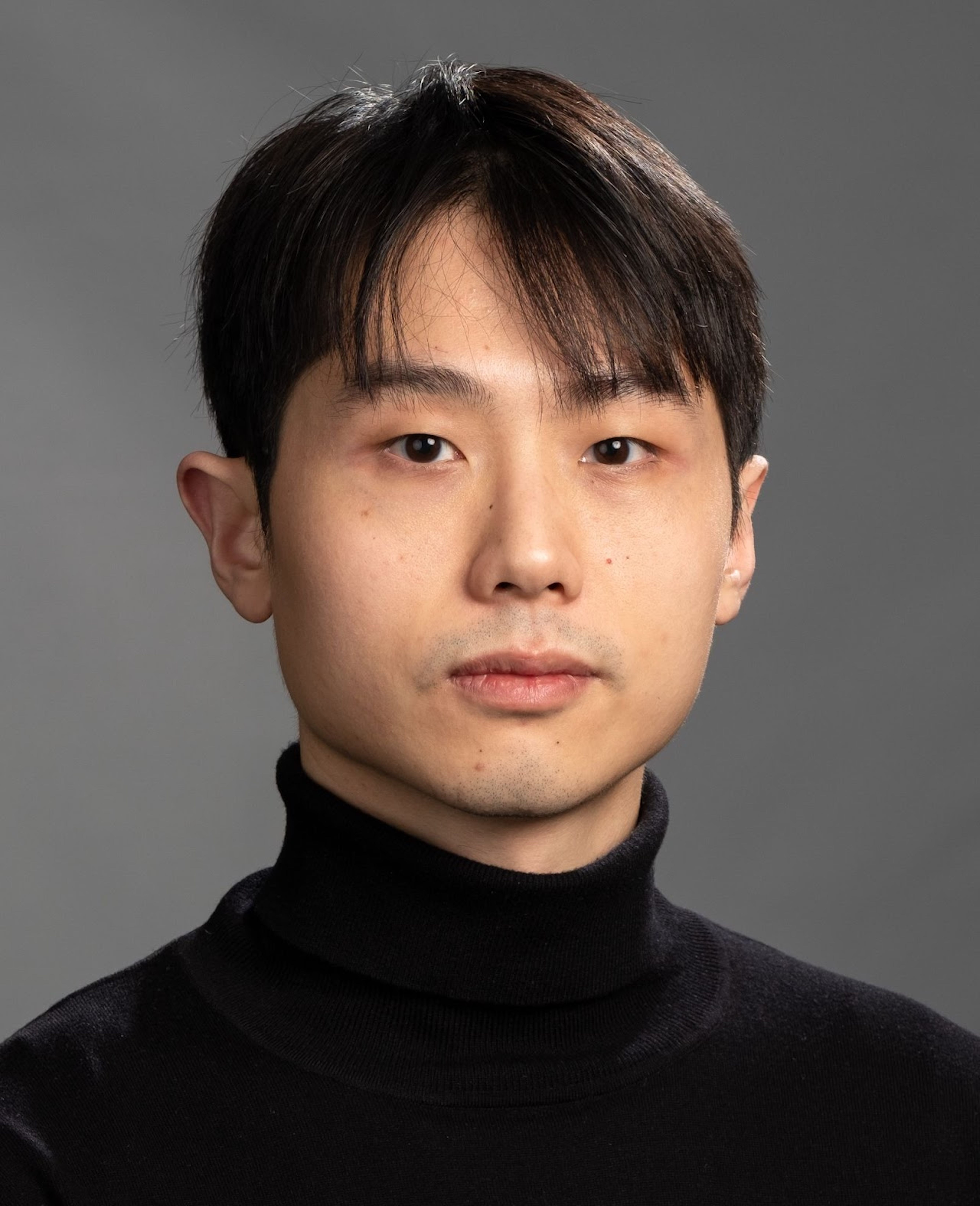}}]{Deng Li}
Deng Li is currently pursuing a Ph.D. degree in the Computational Engineering Department at LUT University (Lappeenranta-Lahti University of Technology LUT), Lappeenranta, Finland. His research interests are in the areas of micro gesture recognition, emotion understanding, document binarization, and self-supervised learning.
\end{IEEEbiography}

\begin{IEEEbiography}[{\includegraphics[width=1in,height=1.25in,clip,keepaspectratio]{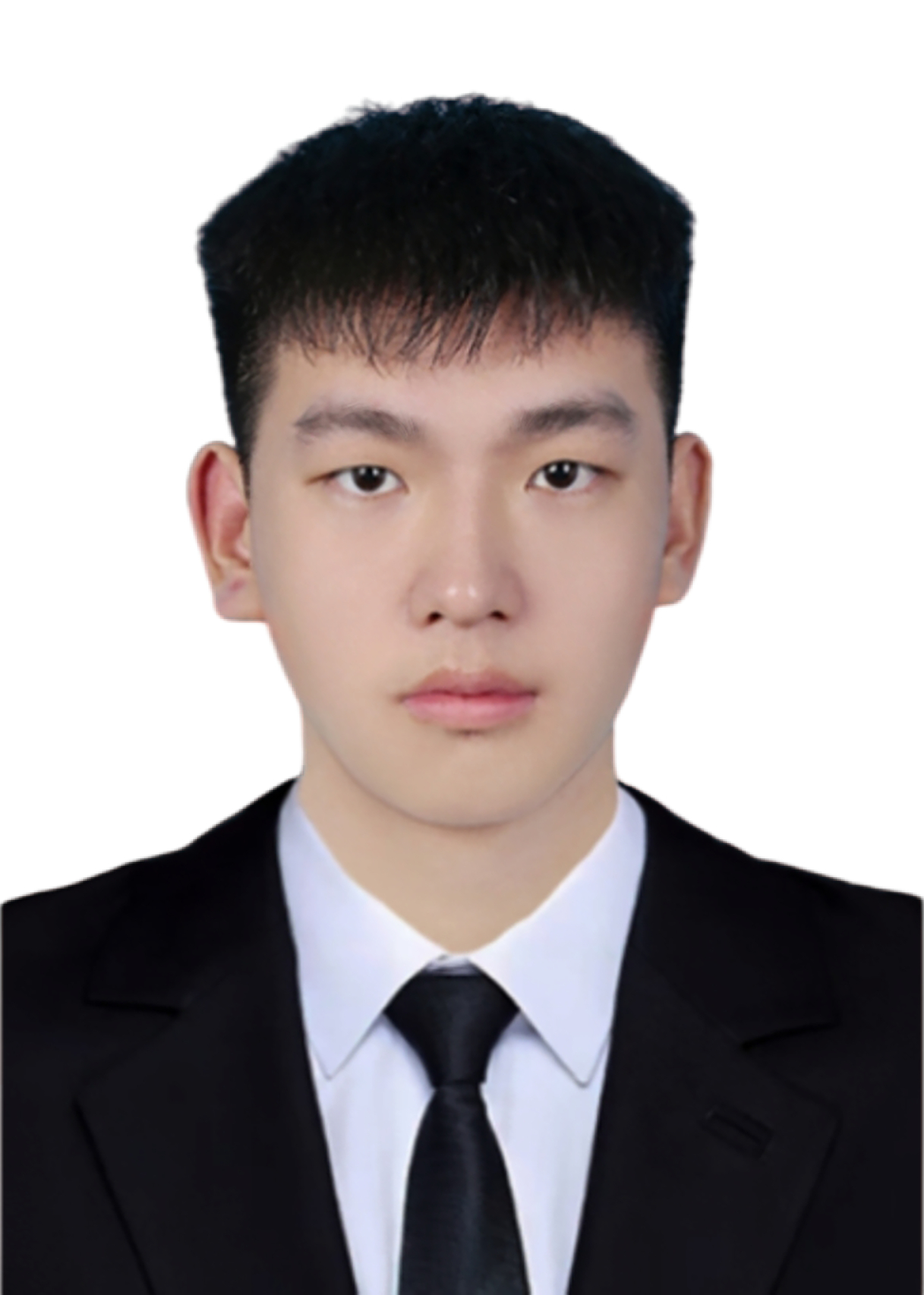}}]{Jun Shao}
Jun Shao received the B.Eng. degree in Communication Engineering from Tianjin University, China, in 2025. He is currently pursuing a M.Eng. degree at Tianjin University, China. His research interests include computer vision and pattern recognition.

\end{IEEEbiography}

\begin{IEEEbiography}[{\includegraphics[width=1in,height=1.25in,clip,keepaspectratio]{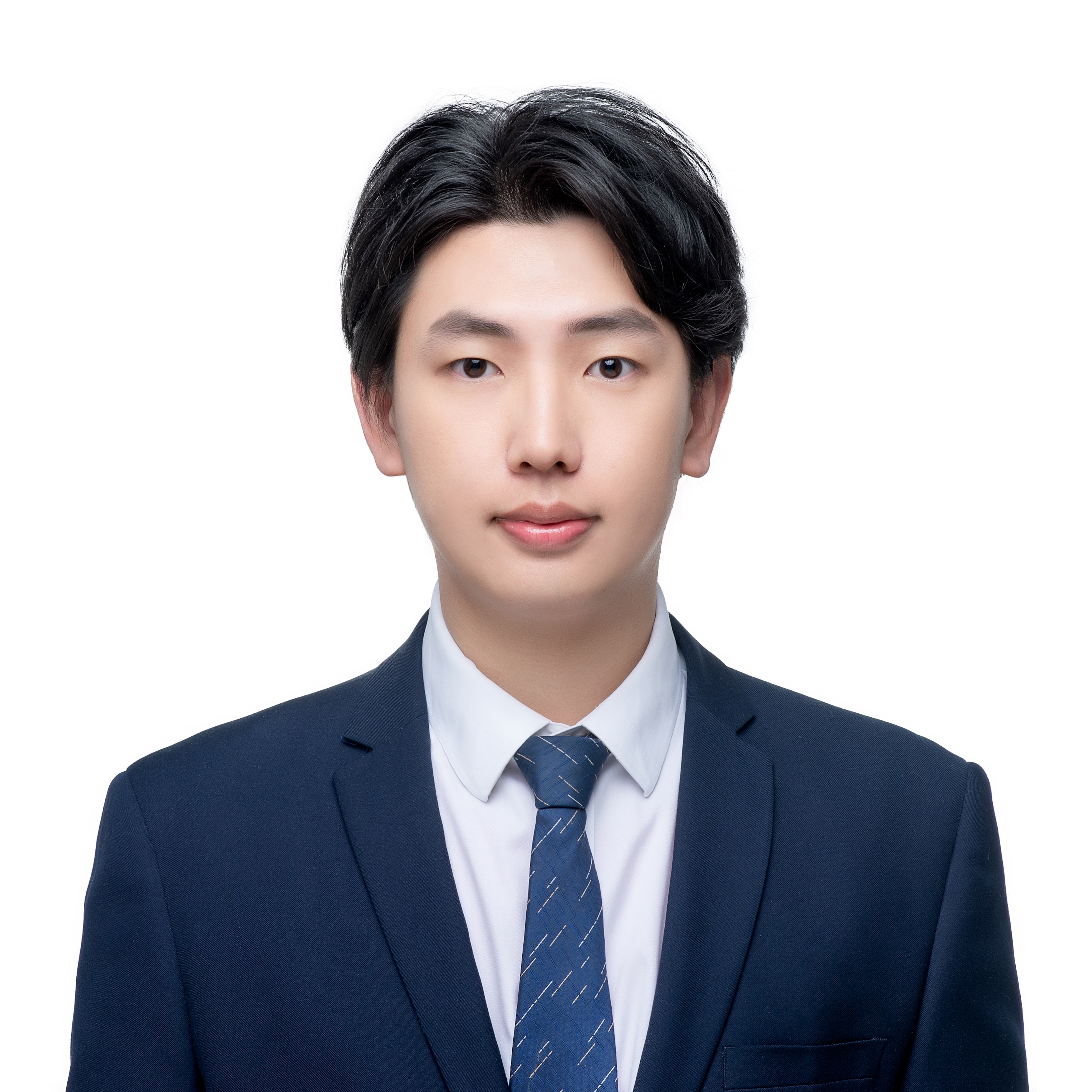}}]{Bohao Xing}
Bohao Xing received the B.Eng. degree in Communication Engineering and the M.Eng. degree in Information and Communication Engineering from Tianjin University, China, in 2021 and 2024, respectively. He is currently a Ph.D. student at Lappeenranta-Lahti University of Technology LUT, Finland. His research interests include affective understanding and emotion AI.
\end{IEEEbiography}

\begin{IEEEbiography}[{\includegraphics[width=1in,height=1.25in,clip,keepaspectratio]{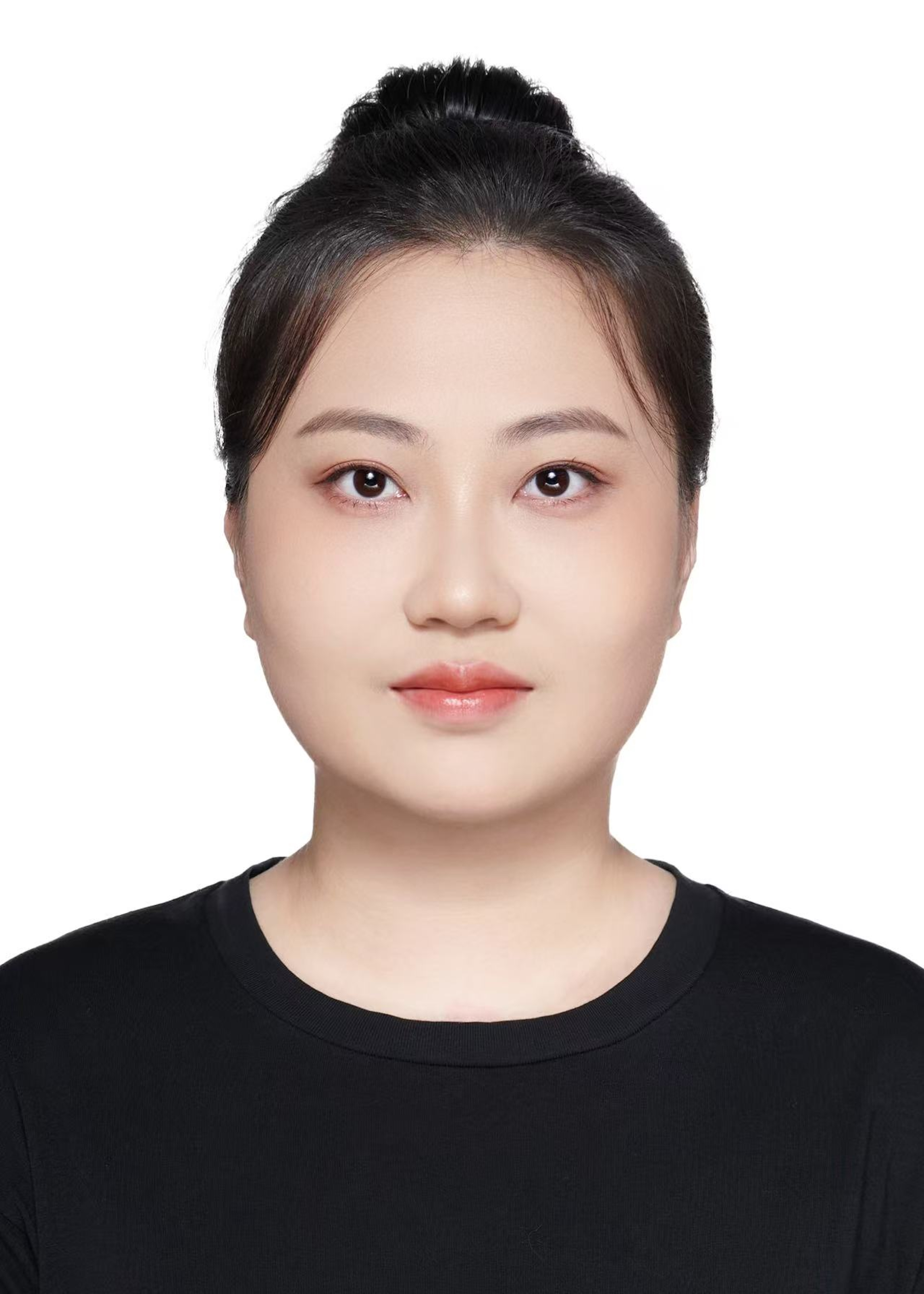}}]{Rong Gao}
Rong Gao received the B.E. degree from Central South University, Changsha, China, in 2020, and the M.E. degree from Tianjin University, Tianjin, China, in 2023. She is currently a Ph.D. student at Lappeenranta-Lahti University of Technology LUT, Finland. Her research interests include computer vision, affective computing, and machine learning.
\end{IEEEbiography}

\begin{IEEEbiography}[{\includegraphics[width=1in,height=1.25in,clip,keepaspectratio]{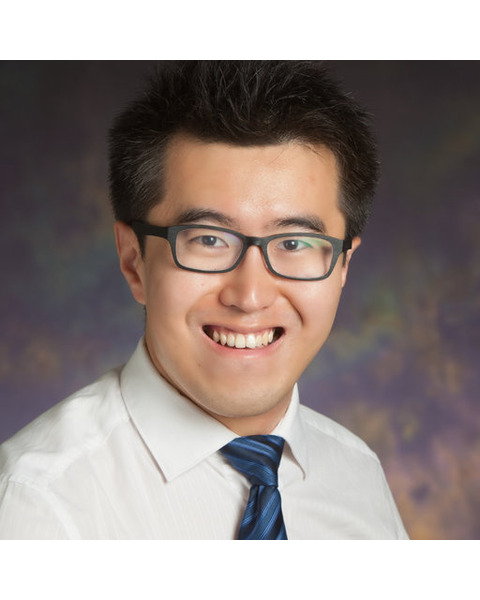}}]
{Bihan Wen}
Bihan Wen (Senior Member, IEEE) received the
BEng degree in electrical and electronic engineer-
ing from Nanyang Technological University (NTU),
Singapore, in 2012, and the MS and PhD degrees in
electrical and computer engineering from the Uni-
versity of Illinois at Urbana–Champaign (UIUC),
Champaign, IL, USA, in 2015 and 2018, respectively.
He is currently a Nanyang assistant professor with
the School of Electrical and Electronic Engineering,
NTU. He received the 2022 Early Career Teaching
Excellence Award and the 2021 Inspirational Mentor
for Koh Boon Hwee Award from NTU, and the 2012 Professional Engineers
Board Gold Medal from Singapore. He was a recipient of the Best Paper Runner
Up Award at the IEEE ICME 2020, the Best Paper Award from IEEE ICIEA
2023, and the Best Paper Award from IEEE MIPR 2023. He was ranked the
World Top 2
consecutively by Stanford University. He was awarded the 2023 CASS VSPC
Rising Star (Runner-Up). He is an associate editor for IEEE Transactions on
Circuits and Systems for Video Technology. He is serving as a guest editor for
IEEE Signal Processing Magazine and IEEE Journal of Selected Topics in Signal
Processing. His research interests include machine learning, computational
imaging, computer vision, image processing, and artificial intelligence security.
\end{IEEEbiography}

\begin{IEEEbiography}[{\includegraphics[width=1in,height=1.25in,clip,keepaspectratio]{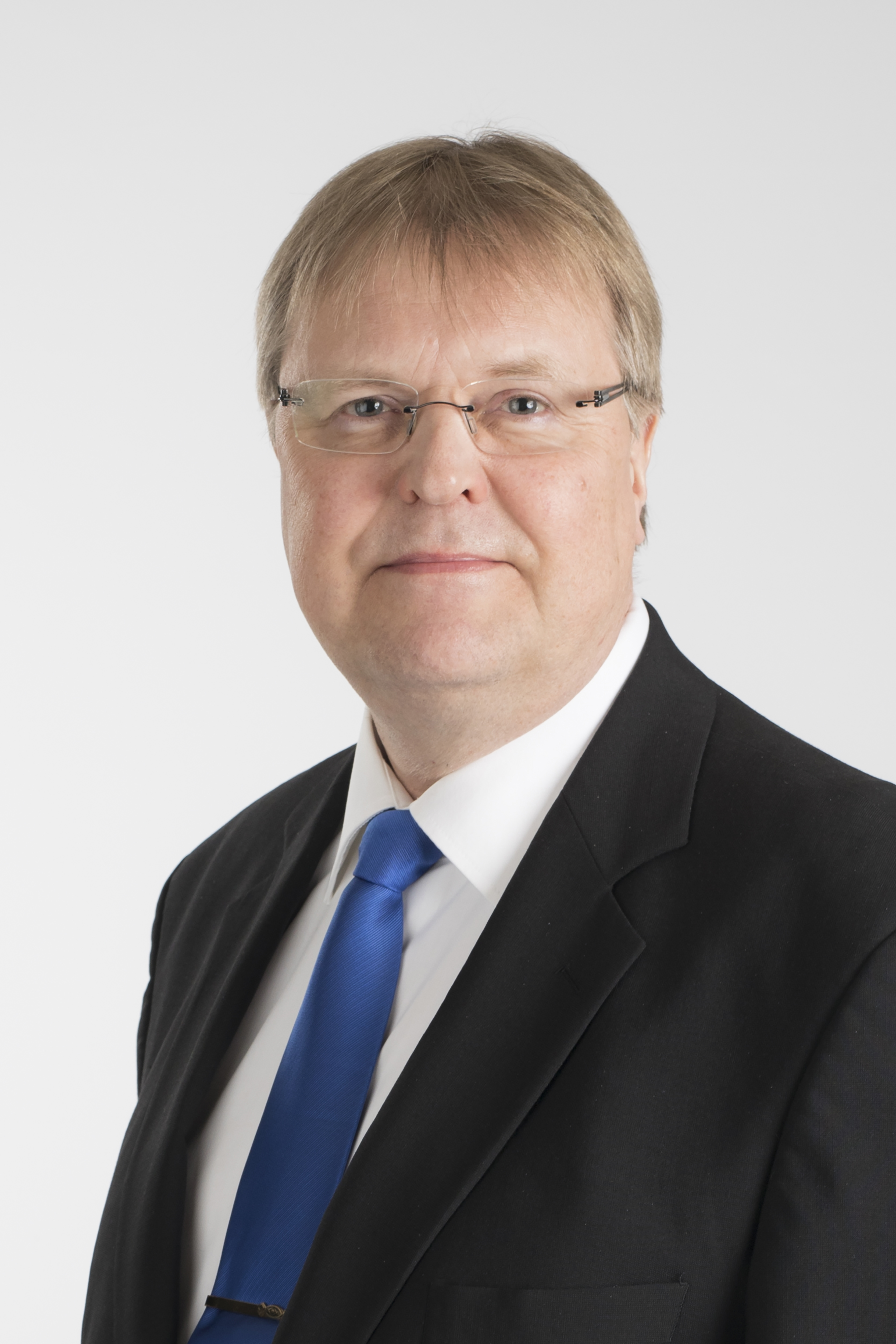}}]{Heikki Kälviäinen}
Heikki Kälviäinen (Senior Member, IEEE) has served as a Professor of Computer Science and Engineering at Lappeenranta-Lahti University of Technology LUT, Finland since 1999 and works in the Computer Vision and Pattern Recognition Laboratory (CVPRL). Besides LUT, since 2025 he has been a Professor in Brno University of Technology, Czech Republic. Prof. Kälviäinen’s research interests include computer vision, pattern recognition, machine learning, and applications of digital image processing and analysis: 30 doctoral theses and 176 master's theses supervised, 37 doctoral dissertations evaluated, more than 230 peer-reviewed scientific articles, and research projects of 6,3M EUR in total. Besides LUT, Prof. Kälviäinen has worked for more than six years in total in the following universities: Brno University of Technology, Czech Technical University, Monash University, Rensselaer Polytechnic Institute, and University of Surrey.
\end{IEEEbiography}

\begin{IEEEbiography}[{\includegraphics[width=1in,height=1.25in,clip,keepaspectratio]{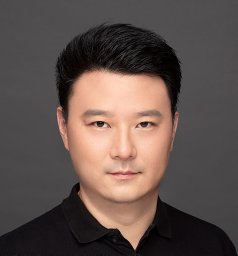}}]
{Xin Liu}
Xin Liu (Senior Member, IEEE) received the Ph.D. degree in computer science and engineering from the University of Oulu, Finland, in 2019. He was at The University of Sydney, as an Endeavour Research Fellow, from 2017 to 2018. He was a Visiting Scholar at the Department of Computing, Imperial College London, in 2019. He held a post-doctoral fellow position at the Academy of Finland. He published more than 80 papers in prestigious journals and conferences, e.g., IEEE Transactions and CVPR, as the first author. Notable accolades include the ICME 2017 Best Paper Award and Runner-Up of ECCV 2020 VIPriors Challenges. His research interests include human behavior analysis, affective computing, social signal processing, emotion AI, and biometrics.
\end{IEEEbiography}

\end{document}